\pdfoutput=1

\documentclass[11pt]{article}

\usepackage[final]{acl}

\usepackage{times}
\usepackage{latexsym}
\usepackage{longtable}

\usepackage[T1]{fontenc}

\usepackage[utf8]{inputenc}

\usepackage{microtype}

\usepackage{inconsolata}

\usepackage{graphicx}

\usepackage{amsmath,amsfonts,bm}
\usepackage{booktabs}
\usepackage{comment}
\usepackage{lipsum}
\usepackage{lineno}
\usepackage{listings}
\usepackage{colortbl}
\usepackage{makecell}
\usepackage{multirow}
\usepackage{supertabular}
\usepackage{authblk}
\usepackage{amssymb}
\usepackage{amsthm}
\usepackage{trajan}
\usepackage{linguex}
\usepackage{fancyvrb}
\usepackage{spverbatim}
\usepackage{enumitem}
\setlist[description]{font=\normalfont\itshape\space}

\usepackage[most]{tcolorbox} 
\tcbset{
  dialoguebox/.style={
    colback=gray!5,
    colframe=gray!50,
    boxrule=0.4pt,
    arc=3pt,
    left=0pt,
    right=4pt,
    top=2pt,
    bottom=2pt
  }
}

\usepackage{xcolor}
\definecolor{lightgreen}{RGB}{220,255,220}
\definecolor{lightred}{RGB}{255,220,220}

\definecolor{darkblue}{rgb}{0, 0, 0.5}
\hypersetup{colorlinks=true, citecolor=darkblue, linkcolor=darkblue, urlcolor=darkblue}

\usepackage[textsize=tiny]{todonotes}

\definecolor{Green}{RGB}{98,115,19}

\newcommand{\clembench}{\texttt{clembench}}

\theoremstyle{definition}
\newtheorem{definition}{Definition}[section]

\input Konanur.fd

\def\eqref#1{equation~\ref{#1}}

\def\1{\bm{1}}

\DeclareMathAlphabet{\mathsfit}{\encodingdefault}{\sfdefault}{m}{sl}
\SetMathAlphabet{\mathsfit}{bold}{\encodingdefault}{\sfdefault}{bx}{n}

\title{{
\fontfamily{pbk}\selectfont
\textcolor{blue}{\rotatebox[origin=c]{15}{\fbox{P}}}
\textcolor{red}{\fbox{L}}
\textcolor{brown}{\fbox{A}}
\textcolor{pink}{\fbox{Y}}
\textcolor{orange}{\rotatebox[origin=c]{-10}{\fbox{P}}}
\textcolor{olive}{\rotatebox[origin=c]{0}{\fbox{E}}}
\textcolor{purple}{\fbox{N}}}\hspace{0.15em}:
An Environment for Exploring\vspace{0.1em}\\ Learning From Dialogue Game Feedback}

\author{
\textbf{Nicola Horst}\textsuperscript{$\otimes$}\thanks{Joint first authorship.},
\textbf{Davide Mazzaccara}\textsuperscript{$\ddagger$}\textsuperscript{$\ast$},
\textbf{Antonia Schmidt}\textsuperscript{$\otimes$}\textsuperscript{$\ast$},
\textbf{Michael Sullivan}\textsuperscript{$\S$},\vspace{0.1em}\\
\textbf{Filippo Momentè}\textsuperscript{$\ddagger$}, 
\textbf{Luca Franceschetti}\textsuperscript{$\triangleleft$},
\textbf{Philipp Sadler}\textsuperscript{$\otimes$},
\textbf{Sherzod Hakimov}\textsuperscript{$\otimes$},\vspace{0.1em}  \\
\textbf{Alberto Testoni}\textsuperscript{$\#$},
\textbf{Raffaella Bernardi}\textsuperscript{$\star$},
\textbf{Raquel Fernández}\textsuperscript{$\triangle$}, 
\textbf{Alexander Koller}\textsuperscript{$\S$}, \vspace{0.1em} \\
\textbf{Oliver Lemon}\textsuperscript{$\circ$},
\textbf{David Schlangen}\textsuperscript{$\otimes$$\oplus$},
\textbf{Mario Giulianelli}\textsuperscript{$\ltimes$},
\textbf{Alessandro Suglia}\textsuperscript{$\diamond$}\vspace{0.1em}
\\

\textsuperscript{$\otimes$}University of Potsdam,
\textsuperscript{$\ddagger$}University of Trento,
\textsuperscript{$\S$}Saarland University, 
\textsuperscript{$\triangleleft$}ETH Zurich,\\
\textsuperscript{$\#$}Amsterdam UMC,
\textsuperscript{$\star$}Free University of Bozen Bolzano, 
\textsuperscript{$\triangle$}University of Amsterdam, \\ 
\textsuperscript{$\circ$}Heriot-Watt University, 
\textsuperscript{$\oplus$}DFKI, 
\textsuperscript{$\ltimes$}UCL,
\textsuperscript{$\diamond$}University of Edinburgh
\vspace{0.1em}\\
\texttt{\href{mailto:lm-playschool@googlegroups.com}{lm-playschool@googlegroups.com}}
}

\begin{document}
\maketitle

\begin{abstract}
Interaction between learner and feedback-giver has come into focus recently for post-training of Large Language Models (LLMs), through the use of reward models that judge the appropriateness of a model's response. 
In this paper, we investigate whether Dialogue Games---goal-directed and rule-governed activities driven predominantly by verbal actions---can also serve as a source of feedback signals for learning.
We introduce \textsc{playpen}, an environment for off- and online learning through Dialogue Game self-play, and investigate a representative set of post-training methods: supervised fine-tuning; direct alignment (DPO); and reinforcement learning with Group Relative Policy Optimization (GRPO). We experiment with post-training a small LLM (Llama-3.1-8B-Instruct), evaluating performance on unseen instances of training games as well as unseen games, and on standard benchmarks. We find that imitation learning through SFT improves performance on unseen instances, but negatively impacts other skills, while interactive learning with GRPO  shows balanced improvements without loss of skills. We release the framework and the baseline training setups to foster research in this promising new direction of ``learning in (synthetic) interaction''.

\vspace{.11em}
\hspace{.15em}\raisebox{-0.2\height}{\includegraphics[width=1em,height=1em]{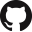}}%
\hspace{.4em}%
\parbox{\dimexpr\linewidth-2\fboxsep-2\fboxrule}{\small \url{https://github.com/lm-playpen/playpen}}

\end{abstract}

\begin{figure}[ht]
\small
\begin{tcolorbox}[dialoguebox]
\begin{description}[noitemsep]
   \item[A (quietly reads game card):] \texttt{Target word: ugly. Taboo words: displeasing, despicable, unattractive}.
   \item[(1a) A (to B):] Something that is not pleasing to the eye, often making a bad impression.
   \item[(1b) B:] Eyesore
   \item[(2a) A:] More about the quality of a person or thing, rather than a nuisance or distraction.
   \item[(2b) B:] Ugly.
   \item[(3a) A:] Yay! Well done us. 
\end{description}
\end{tcolorbox}
\vspace{-0.4em}
\caption{A successful episode of the two-player Dialogue Game \texttt{Taboo}, as played by humans.}
\label{fig:taboo-hh}
\end{figure}

\section{Introduction}

Dialogue Games of the kind illustrated in Figure~\ref{fig:taboo-hh} have been used to evaluate interactive reasoning capabilities of LLMs, requiring as they do the strategic acquisition of information through multi-turn interaction  \cite{bertolazzi-etal-2023-chatgpts,Chalamalasetti-2023,qiao2023gameeval,li2023static,guertler_textarena_2025,cui2025tales}.
In this paper, we show that it is also possible to derive a learning signal from such games, which we call  Dialogue Game Feedback (DGF). DGF offers a unique combination of properties:
A)~Unlike feedback from the reward models used in Reinforcement Learning from Human Feedback \citep[RLHF;][]{ouyang2022rlhf}, DGF is \textit{teleological} or goal-directed, in that it judges not just the appropriateness of a response, but also its effectiveness in advancing toward a desired outcome. 
This is a property DGF shares with feedback from process- and outcome-based reward models used to optimize reasoning models (PRM, \citealt{setlur_rewarding_2024}; ORM, \citealt{hosseini_v-star_2024,cobbe_training_2021}; respectively).
B)~Unlike these aforementioned prior methods, DGF is \textit{objective}, in that it can be computed programmatically, rather than needing a learned, ``subjective'' judgement model;
a property it shares with the ``verifiable rewards'' of \citet{lambert_tulu_2025}. 
C)~Unlike all of these other methods, it can be derived from \textit{inter-subjective}, multi-turn linguistic interaction.
As DGF is defined in terms of \textit{task success}, and task success here is conditional on \textit{communicative success}---i.e., players are required to produce mutually intelligible language---the feedback signal implicitly carries linguistic normative pressure.
D)~Lastly, where other methods focus on alignment with respect to \textit{desirability} (RLHF) or specifically on reasoning skills in domains such as maths and coding (PRM, ORM, verifiable rewards), DGF rewards general backwards- and forwards-looking language use. At the same time, insofar as the underlying game requires them, DGF also targets specific skills such as spatial reasoning or referential language.\looseness-1

Figure~\ref{fig:taboo-hh} illustrates this. It shows an example of two players playing the Dialogue Game \texttt{Taboo} where a \textit{clue giver} needs to describe a concept to a \textit{guesser}, while avoiding certain ``taboo'' words. Both players produce natural language strings, which can be judged in two ways: \textit{teleologically}, where both 1a and 1b are appropriate but do not lead to success, while 2a and 2b succeed; and \textit{objectively} as it can be programmatically verified that 1b and 2b comply with the rules and that 2b is the correct answer.
Moreover, for the game to progress, both clue giver and guesser must produce mutually intelligible language, placed in the context of the interaction as a whole. E.g.\ turn 2a  must consider the previous guess and how it failed (backwards-looking) in order to produce an utterance that aims to elicit a better guess (forward-looking).

\begin{figure*}[ht!]
    \centering
    \includegraphics[width=.9\linewidth]{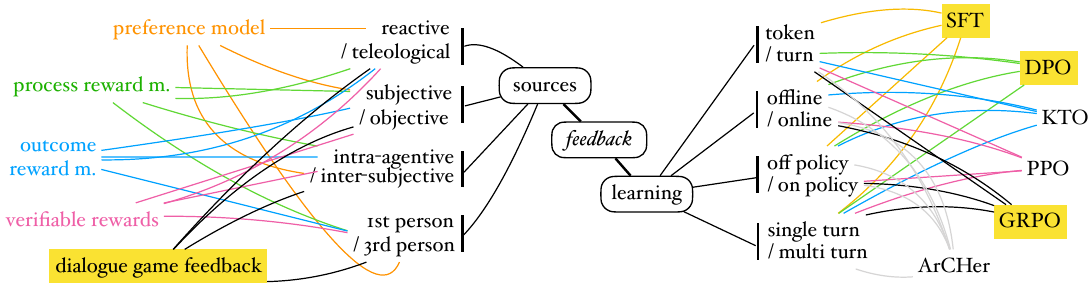}
    \caption{Placing Dialogue Game Feedback in the wider research context. Highlighted on the right the selection of learning methods that we provide baselines for.}
    \label{fig:schematic}
\end{figure*}


\ \\
\noindent
\textbf{Contribution 1:} We introduce \textsc{playpen}, an environment where LLMs engage in dialogue games (without human intervention) from which a feedback signal for learning is derived; either online during gameplay or offline from rollouts, and based either on single trajectories or sets of ranked alternatives.
\textbf{Contribution 2:} 
We demonstrate how to leverage this feedback signal for post-training.\footnote{%
    Leaving to future work the exploration of these types of interactive settings for language acquisition from scratch.
}
We investigate three learning approaches: imitation learning (supervised fine-tuning); an offline alignment approach, Direct Preference Optimization \citep[DPO;][]{rafailov2023direct}; and an online learning algorithm, Group Relative Policy Optimization \citep[GRPO;][]{shao2024deepseekmath}---thereby establishing strong baselines for this learning environment.
We evaluate the resulting models on a range of tests, including held-out dialogue games to assess skill generalisation, a comprehensive suite of tests assessing broader linguistic and cognitive abilities \citep{momente2025triangulating}, and standard NLP benchmarks such as MMLU-Pro \citep{mmlupro2024}, Big-bench Hard \citep{suzgun-etal-2023-challenging}, and IFEval \citep{ifeval}. 
\textbf{Contribution~3:} We show that imitation learning through SFT improves performance on unseen instances but negatively impacts other skills, whereas interactive learning with GRPO achieves balanced improvements without skill degradation.




\section{Related Work}
\label{sec:related}


Our work builds on several threads of research, which we briefly review here and visualise in Fig.~\ref{fig:schematic}.

\paragraph{Sources of Learning Feedback.}

Post-training methods typically assume some \textit{feedback} on the appropriateness of a model's production $R$ in a context $C$, via a feedback function $f$.
We outlined in the introduction the main approaches to defining $f$, from trained judge models to verifiers in formal domains \citep[see][for recent overviews]{kumar_llm_2025,rlhf2024}, and positioned our \textit{Dialogue Game Feedback} within this landscape.
The feedback function either judges $(C,R)$---we may call this \textit{direct feedback}---or ranks candidate responses $(C, \{R, R', \dots\})$ in a \textit{comparative judgement}.
Most methods judge the \textit{whole} production $R$; but see \citet{wu_fine-grained_2023} for more fine-grained proposals.
We can make one further distinction. 
The approaches discussed so far can be seen as providing \textit{outside} or \textit{3rd person feedback}, in that the judge observes but does not participate in the interaction.  
Others \citep[e.g.,][]{sumers2021learning,chen2024retrospective} extract evidence from the ongoing interaction itself, or inject such feedback synthetically \citep{akyurek-etal-2023-rl4f,wangmint}---what we call \textit{inside} or \textit{1st person} feedback.\footnote{There is a much older parallel line on feedback in language learning, from Luc Steels' work \citep{steels_evolving_2003,nevens2017computational} to more recent `emergent communication' studies \citep[e.g.,][]{Cao2018}. How to adapt this for LLM post-training remains open.} 
It has been variously explored (see e.g.\ \citealt{sumers2021learning,chen2024retrospective}) to take evidence from the ongoing interaction as well.
Our framework can incorporate such ``inside" feedback, but here we focus on what can be learned from ``outside"---programmatically computed game feedback.  
We regard DGF as both \textit{teleological} (goal-directed) and \textit{objective}, in that $f$ is defined algorithmically and encodes the game’s specific goals. 
A similar kind of feedback has been used recently by \citet{gul_cogen_2024}, but only in the context of a single game, and with specialised learning methods. Similarly, \citet{Sadler-2024-1} used feedback from a cognitively inspired programmatic partner to bootstrap the language capabilities of a collaborative neural agent in a single visual reference task.








\paragraph{Methods for Learning from Feedback.}

Assuming that a feedback function $f$ is at hand, how can this be used to improve the model's production policy? Various methods have been developed in recent years to do this (see the surveys cited above). 
A first way of categorising them is according to their granularity.
\textit{Token-based} methods, such as supervised fine-tuning, essentially use $f$ to filter data for imitation learning.
The hypothesis is that producing sequences like those observed during training generalises to improving the model's strategic behaviour.
\textit{Turn-level} methods instead make direct use of the turn-level feedback obtained from $f$. 
Within these, \textit{offline} variants such as Direct Preference Optimization \citep[DPO;][]{rafailov2023direct} and Kahneman-Tversky Optimization \citep[KTO;][]{ethayarajh2024kto} train on preference judgements from previously collected data, which do not need to come from the current policy (in this sense, they are both offline and off-policy).
On the other hand, \textit{online} methods such as Proximal Policy Optimization \citep[PPO;][]{schulman2017proximal} and Group Relative Policy Optimization \citep[GRPO;][]{shao2024deepseekmath} attempt to directly improve the policy that produces the samples.
Finally, multi-turn methods such as ArCHer \citep{zhou_archer_2024}, ReSpect \citep{chen2024retrospective}, and REFUEL \citep{gao2024regressing} are emerging, which explicitly handle the turn-level structure of conversations; we leave these for future work and here focus on a representative selection of methods: SFT, DPO, and GRPO.


\paragraph{LLMs and Dialogue Games.} 

Conversational interactions framed as games have long been used to investigate language use; see discussion in  \citep{Schlangen-2019-1,Schlangen-2023-1,Suglia2024}, which also proposed to use them for evaluating language use capabilities of NLP models. This idea has been implemented by various frameworks in recent years. 
Early frameworks such as TextWorld \citep{cote2019textworld} supported only a single game genre (interactive fiction) and assumed specialised models. 
Only with the advent of generalist models that can be \textit{prompted} into being specialists \cite{brown_language_2020,wei_finetuned_2021} did it become possible to implement this idea at a larger scale, both for single games \cite{bertolazzi-etal-2023-chatgpts, mazzaccara-etal-2024-learning} and in multi-game frameworks \cite{Chalamalasetti-2023,qiao2023gameeval,li2023static,gong2023mindagent,wu2024smartplay,Zhou2024-sotopia,duan2024gtbench,guertler_textarena_2025,cui2025tales}.
Among these, we chose to build on \clembench\ \cite{Chalamalasetti-2023} as the longest-running continuously maintained effort, which also comes with an extensive archive of dialogue game transcripts spanning a wide range of both open and closed models.

\paragraph{The Cognitive Plausibility of Learning from Interaction.}

It is well established in the developmental literature that human language acquisition requires social interaction \cite{clark_first_2016,bruner_childs_1983,kuhl_is_2007,hiller-fernandez-2016-data,saxton2000negative,bloom_how_2000}, and similar arguments have been made for machine language learning \cite{fernandez-EtAl:2011:IJCAI,Bisk2020,Bender2020}, especially given stark differences in sample efficiency between human and artificial learners \citep{hart-risley-1995,cristia2019child,linzen-2020-accelerate}. 
Our work speaks to this hypothesis by exploring a learning signal derived from (an approximation of) social linguistic interaction.
Incidentally, this is also the motivation of the BabyLM challenge \citep{charpentier_babylm_2025}, which in its latest incarnation explicitly encourages the use of synthetic interaction. 

A few recent pioneering works \citep{nikolaus-fourtassi-2021-modeling,ma_babysit_2025} 
have begun to explore this direction, demonstrating the potential benefits of interaction even in learning from scratch.
In this work, our focus is on post-training, and we assume that models can already follow instructions well enough to engage in gameplay.


\section{Playpen: Dialogue Games \& Feedback}
\label{sec:dgpp}

\subsection{Dialogue Games with LLMs}

Figure~\ref{fig:taboo-hh} above provided an example of a Dialogue Game. How can we enable LLMs to play such games effectively? One of the surprising insights of the ``LLM revolution'' was that, at previously unseen scale, these models can be \textit{prompted} to perform a wide range of tasks \cite{wei_finetuned_2021,brown_language_2020}. As the frameworks described above have shown, this extends to prompting LLMs to act as policies for playing conversational games---albeit with varying degrees of success. For example, a simple prompt such as ``\textit{We are playing a word guessing game. Your task is to describe the word, but you are not allowed to use some other words. The word to describe is `ugly', and the words to avoid are `displeasing', `despicable', and `unattractive}'.'' can induce (at least some) LLMs to act as a policy $\pi_{taboo}$ capable of playing that specific role in the game reasonably well.\footnote{%
    See \citet{Chalamalasetti-2023} for an example of a full prompt that can be used, which needs to contain additional formatting instructions.
}
To enable self-play with LLMs, each player must be separately prompted, often with distinct information states. Following \citet{Chalamalasetti-2023,smith_concordia_2024}, we use a programmatic \textit{Game Master} (GM) to mediate the interaction. In the case of Figure~\ref{fig:taboo-hh}, for example, the GM would insert a turn between 1a and 1b, delivering instructions to player B and relaying the clue from player A. See Appendix~\ref{app:dialogue_games_transcripts} for full transcripts of such interactions.

All of the games used here (see Section \ref{subsec:playpen-env} below) involve some form of reasoning. Crucially, however, the reasoning involved is fundamentally different from that required in standard applications of reasoning models \cite{besta_reasoning_2025}. Unlike conventional reasoning tasks such as math word problems \cite{hendrycks_measuring_2021}, which are \textit{well-posed}---i.e., the problem is fully specified and the challenge lies in deriving a solution through a correct sequence of steps---the games studied here require \textit{multi-turn} and \textit{interactive} reasoning, as they are \textit{ill-posed} at the outset. 
They only become tractable through iterative exchanges between players. Consider the starting prompt for a \texttt{Wordle}-type word guessing game: ``\textit{Guess a 5 letter word}''. Only through making guesses, receiving feedback, and updating beliefs accordingly does the task become solvable and the identity of the target word recoverable. The reasoning at play in such settings involves managing uncertainty and coordinating with another agent under conditions of imperfect information.\looseness-1



\subsection{Dialogue Game Feedback}
\label{subsec:dialogue-game-feedback}

We refer to games such as \texttt{Taboo} or \texttt{Wordle} as \textit{dialogue games} (DG).
A particular instantiation of a DG---obtained by filling in a prompt template with specific parameters (for example, the target and taboo words in \texttt{Taboo})---is called a \textit{dialogue game realisation instance}, or simply an \textit{instance}. We denote instance $i$ of game $g$ as $x_{g,i}$.
Each instance defines a \textit{game tree}, which originates from the initial game description and branches out at every turn into all possible actions available at that point. In other words, the tree contains all possible gameplays for that particular instance. 
If the DG 
allows verbal actions of unbounded length (i.e., compositional and infinite action spaces), the corresponding game tree will have an infinite number of nodes and edges. A (complete) \textit{trajectory} is a path from the root node to a leaf node. Each player in the game is represented by a \textit{policy} $\pi$ specifying their action at each decision point.
When all players in a game are instantiated with policies, this collectively induces a distribution over trajectories. An \textit{episode} of a given instance can be viewed as a sample from this distribution. The resulting interaction can be recorded as a \textit{transcript} $t$ as follows (this description applies to a two-player game, with straightforward extensions to games involving more players).

\begin{definition}[Transcript]
A \textit{transcript} $t$ represents a trajectory through a game instance tree:
\begin{align*}
    t &= (S_0, C_1^A, R_1^A, S'_{0}, C_1^B, R_1^B, S_1, \dots,\\
      &  S_{F-1}, C_F^A, R_F^A, S'_{F-1}, C_F^B, R_F^B, S_F)
\end{align*}
where $C_n^P$ is the context shown to player $P$ at turn $n$, $R_n^P$ is their response, and $S_n$ is the abstract game state at turn $n$.
The initial $C_1^P$ has a special status, as it contains the game description.
A trajectory is \textit{complete} if it ends in a final state $S_F$ or an abort state $S_X$.  
Dialogue Game Feedback is given by a game-specific feedback function $f$ which evaluates trajectories.
The scoring function typically evaluates complete trajectories, although in certain games---including some of the games described below---incomplete trajectories may also be assessed.\looseness-1 
\end{definition}

A transcript $t$ as defined above contains all interleaved interactions between players and the Game Master.
However, each player $P$ only observes the contexts $C^P$ presented to them, never the raw responses of other players. 
To recover a player's view, a perspective function $p_P$ reduces a trajectory to the sequence of context–response pairs $(C_i^P, R_i^P)$ for $0\leq i < n $ that player $P$ has experienced at turn $n$.\looseness-1

\subsection{The \textsc{playpen} Environment}
\label{subsec:playpen-env}
The \textsc{playpen} environment we introduce here builds upon the Dialogue Game benchmark \cite[\texttt{clembench};][]{Chalamalasetti-2023}, transforming it into an interactive playground in which LLMs can learn to be language users.
At the time of writing, \textsc{playpen} includes 15 \clembench\ Games, testing language and world knowledge (e.g., in games such as \texttt{Taboo}, \texttt{Wordle}, \texttt{Codenames}), the ability to perform conversational grounding (e.g.\ \texttt{PrivateShared}, \texttt{GuessWhat}), and spatial and causal reasoning (e.g., \texttt{Adventure} \texttt{Games} or \texttt{Map} \texttt{Navigation}). We provide the full list of games with further details in Appendix~\ref{app:games}.

By recording trajectories as defined in Section \ref{subsec:dialogue-game-feedback}, \textsc{playpen} supports both offline and online learning, as well as the representation of branching subtrees within the overall game tree through repeatedly sampling from player policies. This flexibility enables the learning experiments that we turn to now.

\section{Experimental Setup}
\label{sec:setup}
Our experiments focus on leveraging Dialogue Game Feedback to post-train language models, building on the premise that this feedback signal is most effectively used when the model is already capable of prompted gameplay to a certain extent.

\subsection{Models} 

We selected Llama-3.1-8B and Llama-3.1-70B \citep{meta_llama_2024}, both in the \texttt{Instruct} variant, as they have generally shown to be receptive to further training \citep{taori2023alpaca, zhou2023lima}, and have performed well within their size classes on the public \clembench\ leaderboard.\footnote{
    \url{https://clembench.github.io/leaderboard.html}, accessed 2025-05-09.
}
We also report results for Qwen2-7B \citep{team2024qwen2} (also in its \texttt{Instruct} variant) to showcase the effect of the different training regimes on another state-of-the-art LLM as well.
In some experiments, we also used a 4-bit quantised version of the model for more efficient training and inference.\footnote{\url{https://huggingface.co/unsloth/Meta-Llama-3.1-8B-Instruct-bnb-4bit}}




\subsection{Evaluation}
\label{subsec:setup-evaluation}
What improvements can we expect from learning with Dialogue Game Feedback? We hypothesise that we will see improvements in gameplay on unseen instances of the games encountered during training, as well as generalisation to new game types. To assess the broader impact of DFG learning, we additionally evaluate the post-trained models for their formal and functional linguistic competence, as well as on general NLP benchmarks.

\vspace*{-.15cm}
\paragraph{Interactive Dialogue Games.} 
Performance on interactive dialogue games is evaluated using the \texttt{clemscore} metric \citep{Chalamalasetti-2023}, which captures both the ability to adhere to the formal rules of a game and the quality of the gameplay.
Specifically, the \texttt{clemscore} is obtained by multiplying the macro-average percentage of games that were validly played with the macro-average quality score (typically task success) in those valid attempts.
We use 7 of the 15 available Dialogue Games 
for training: \texttt{Taboo}, \texttt{PrivateShared}, \texttt{ImageGame}, \texttt{ReferenceGame}, and three variants of \texttt{Wordle}. 
For evaluation, we generate new instances of these games to form the \textit{in-domain} test set. 
The remaining eight games---\texttt{Codenames}, \texttt{Adventure} \texttt{Game}, \texttt{GuessWhat}, \texttt{MatchIt}, and three variants of \texttt{Map} \texttt{Navigation}---serve as an additional \textit{out-of-domain} test set.


\vspace*{-.15cm}
\paragraph{Formal Linguistic Competence.}
We evaluate formal linguistic competence \citep{mahowald2024dissociating}, such as the ability to recognise morphosyntactic agreement or lexical entailment, using the GLUE Diagnostic dataset \citep{wang2018glue}.

\vspace*{-.15cm}
\paragraph{Functional Linguistic Competence.} 
We evaluate cognitive abilities required for verbal interaction, such as working memory, common-sense reasoning, theory of mind and other executive and socio-emotional skills, using a subset of the tasks curated by \citet{momente2025triangulating}.
Specifically, we use a sample from Natural Plan \citep{zheng2024naturalplanbenchmarkingllms}, LogiQA 2.0 \citep{logiqa}, CLadder \citep{cladder}, WinoGrande \citep{winogrande}, EQ-Bench \citep{eqbench}, LM-Pragmatics \citep{hu-etal-2023-fine}, SocialIQA \citep{sap-etal-2019-social}, and SimpleToM \citep{simpletom}.\looseness-1 


\vspace*{-.15cm}
\paragraph{Knowledge and Instruction Following.}
Finally, we evaluate models on two widely used LLM benchmarks: 
MMLU-Pro \citep{mmlupro2024} and Big-bench Hard \citep[BBH;][]{suzgun-etal-2023-challenging,srivastava2023beyond}. In addition, we report performance on IFEval \citep{ifeval}, which assesses general instruction-following capabilities.

\subsection{Training Regimes and Interaction Data}
\label{subsec:setup-training-regimes}
We experiment with several training regimes enabled by \textsc{playpen}, and compare their outcomes with the unmodified Llama-3.1 (\texttt{Baseline}). 
Additional details on training setups, hyperparameters, and data generation procedures are in Appendix~\ref{app:training_details}. 

\subsubsection{Imitation Learning (SFT)}
\label{subsubsec:imitation_learning}

We begin by investigating the potential of pure imitation learning through supervised fine-tuning (SFT). 
We create a training dataset $\mathcal{D}$ by collecting episodes of gameplay from a large variety of models listed on the public \clembench\ leaderboard.
For our experiments with SFT, we filter $\mathcal{D}$ to retain only the 7079 successful episodes (i.e., we discard lost and aborted episodes) and convert the transcripts into separate trajectories for all player perspectives:
$\mathcal{D}_{SFT} = \{p_A(t), p_B(t) \;|\; t\in\mathcal{D} ,\, s(t) > \tau_g\}$, for a game-specific threshold $\tau_g$, and using player-specific perspective functions $p_A$ and $p_B$ (see Section~\ref{subsec:dialogue-game-feedback}).
We fine-tune the \texttt{Baseline} model using different data mixtures, containing only interaction data or also instruction-following examples. 

After initial experimentation with plain SFT on the 7079 transcripts over 1100 training steps,\footnote{Details on this analysis are reported in Appendix~\ref{appendix:training_details_sft}.} we found that more sample-efficient variants offered better generalisation.  
In what follows, we report results from the most effective configuration, which we refer to as \texttt{SFT (Cold Start, CS)}. This variant uses only 700 training steps and focuses exclusively on interaction data.

\subsubsection{Direct Alignment}
\label{subsubsec:direct_alignment}
We apply Direct Preference Optimization \citep[DPO;][]{rafailov2023direct} to the best SFT checkpoint, \texttt{SFT (CS)};
DPO offers a middle ground between SFT and full online reinforcement learning by leveraging contrastive learning on offline data.
We consider two variants of DPO training, using dialogue-level or turn-level preference pairs, respectively.

\paragraph{\texttt{DPO Dialogue}.} For every positive sample in the filtered dataset $\mathcal{D}_{SFT}$, we find a negative counterpart in $\mathcal{D}$ that starts from the same initial state (prompt and game instance $S_0$) but ends without reaching a successful final state. This includes both aborted and completed but unsuccessful dialogues. The dataset consists of ca.\ 10K pairs of positive and negative trajectories.
    
\paragraph{\texttt{DPO Turn}.} For each turn in a successful dialogue, we find a negative counterpart that shares the same conversational history (the prompt, the game instance and the history up to a branching point) to yield $S_0, C^A_1, R^A_1, \ldots C^A_n, (R_n^A, R'^A_n)$, i.e., paired samples identical up to the penultimate branching node in the game tree. 
The dataset consists of ca.~86K pairs of positive and negative trajectories.

\subsubsection{Online Learning}
While both SFT and DPO can provide useful learning signals for gameplay, neither method captures the interactive nature of dialogue games. 
For this reason, we also performed experiments using GRPO~\citep{shao2024deepseekmath}.
Unlike the above methods, GRPO does not rely on a fixed, offline (and possibly off-policy) training dataset $\mathcal{D}$. 
Instead, for each game instance $x_i$ (initial prompt), we use the target model to interactively produce $8$ samples of full gameplay (with temperature set to $0.75$). This leads to a group $G_i$ of alternative transcripts.
Each trajectory $g \in G_i$ is evaluated using a game-specific reward function that corresponds to the quality score computation for that game in the \clembench\ benchmark (see Appendix \ref{app_grpo_data_gen}).
We test two configurations of GRPO, with training starting either from the base or the \texttt{SFT (CS)} model.

\section{Results}
\label{sec:results}
We now present the results of running our evaluation suite (Section~\ref{subsec:setup-evaluation}) on the post-trained models.
Our main finding is that only GRPO learned from Dialogue Game Feedback in a way that generalised to unseen games, without degrading other skills. 
This is the sole interactive, turn-based method tested.
Other methods appeared to overfit to the training distribution and failed to transfer.

\begin{table*}[ht]
\centering
\resizebox{.95\textwidth}{!}{%
\begin{tabular}{lllllll}
\toprule
 & \multicolumn{3}{c}{\textbf{In Domain}} & \multicolumn{3}{c}{\textbf{Out of Domain}} \\
\cmidrule(lr){2-4} \cmidrule(lr){5-7}
\textbf{Model} & \textbf{Clemscore} $=$ & \textbf{\% Played} $\times$ & \textbf{Quality} & \textbf{Clemscore} $\uparrow$ & \textbf{\% Played} $\uparrow$ & \textbf{Quality} $\uparrow$ \\
\midrule
\textbf{Llama-3.1-8B} &&&&&& \\
Baseline & 19.39 & 58.50 & 33.15 & 24.58 & 54.53 & 45.08 \\
SFT (CS) & 40.11 {\scriptsize \textcolor{green!60!black}{\rule{20.72pt}{5pt}}} & 70.48 {\scriptsize \textcolor{green!60!black}{\rule{11.98pt}{5pt}}} & 56.91 {\scriptsize \textcolor{green!60!black}{\rule{23.76pt}{5pt}}} & 22.53 {\scriptsize \textcolor{red}{\rule{2.05pt}{5pt}}} & 50.55 {\scriptsize \textcolor{red}{\rule{3.98pt}{5pt}}} & 44.58 {\scriptsize \textcolor{red}{\rule{0.05pt}{5pt}}} \\


SFT (CS) + DPO (Dial.) & 32.33 {\scriptsize \textcolor{green!60!black}{\rule{12.94pt}{5pt}}} & 63.54 {\scriptsize \textcolor{green!60!black}{\rule{5.04pt}{5pt}}} & 50.89 {\scriptsize \textcolor{green!60!black}{\rule{17.74pt}{5pt}}} & 19.50 {\scriptsize \textcolor{red}{\rule{5.08pt}{5pt}}} & 51.67 {\scriptsize \textcolor{red}{\rule{2.86pt}{5pt}}} & 37.74 {\scriptsize \textcolor{red}{\rule{7.34pt}{5pt}}} \\

SFT (CS) + DPO (Turn.) & 39.65 {\scriptsize \textcolor{green!60!black}{\rule{20.26pt}{5pt}}} & 71.28 {\scriptsize \textcolor{green!60!black}{\rule{12.78pt}{5pt}}} & 48.90 {\scriptsize \textcolor{green!60!black}{\rule{15.75pt}{5pt}}} & 23.81 {\scriptsize \textcolor{red}{\rule{0.77pt}{5pt}}} & 43.03 {\scriptsize \textcolor{red}{\rule{11.50pt}{5pt}}} & 54.76 {\scriptsize \textcolor{green!60!black}{\rule{9.68pt}{5pt}}} \\

GRPO & 24.89 {\scriptsize \textcolor{green!60!black}{\rule{5.50pt}{5pt}}} & 57.55 {\scriptsize \textcolor{red}{\rule{0.95pt}{5pt}}} & 43.25 {\scriptsize \textcolor{green!60!black}{\rule{10.10pt}{5pt}}} & 33.92 {\scriptsize \textcolor{green!60!black}{\rule{9.34pt}{5pt}}} & 67.38 {\scriptsize \textcolor{green!60!black}{\rule{13.05pt}{5pt}}} & 50.34 {\scriptsize \textcolor{green!60!black}{\rule{5.26pt}{5pt}}} \\

SFT (CS) + GRPO & 24.30 {\scriptsize \textcolor{green!60!black}{\rule{4.91pt}{5pt}}} & 63.22 {\scriptsize \textcolor{green!60!black}{\rule{4.72pt}{5pt}}} & 38.44 {\scriptsize \textcolor{green!60!black}{\rule{5.29pt}{5pt}}} & 31.81 {\scriptsize \textcolor{green!60!black}{\rule{7.23pt}{5pt}}} & 67.26 {\scriptsize \textcolor{green!60!black}{\rule{12.73pt}{5pt}}} & 47.29 {\scriptsize \textcolor{green!60!black}{\rule{2.21pt}{5pt}}} \\
\hline
\textbf{Llama-3.1-70B} &&&&&& \\
Baseline & 37.24 &  64.57 & 57.67 & 47.37 &   82.29 & 57.57 \\
SFT (CS) & 53.60 {\scriptsize \textcolor{green!60!black}{\rule{ 16.37pt}{5pt}}} & 81.57 {\scriptsize \textcolor{green!60!black}{\rule{17.00pt}{5pt}}} & 65.71 {\scriptsize \textcolor{green!60!black}{\rule{8.05pt}{5pt}}} & 54.40 {\scriptsize \textcolor{green!60!black}{\rule{7.03pt}{5pt}}} &  85.57 {\scriptsize \textcolor{green!60!black}{\rule{3.29pt}{5pt}}} & 63.57 {\scriptsize \textcolor{green!60!black}{\rule{6.00pt}{5pt}}} \\
\multicolumn{1}{l}{SFT (CS) + DPO (Dial.)} & 36.92 {\scriptsize \textcolor{red}{\rule{0.32pt}{5pt}}} & 52.44 {\scriptsize \textcolor{red}{\rule{12.13pt}{5pt}}} & 70.41 {\scriptsize \textcolor{green!60!black}{\rule{12.74pt}{5pt}}} & 45.46 {\scriptsize \textcolor{red}{\rule{1.91pt}{5pt}}} & 73.66 {\scriptsize \textcolor{red}{\rule{8.63pt}{5pt}}} & 61.71 {\scriptsize \textcolor{green!60!black}{\rule{4.14pt}{5pt}}} \\
\multicolumn{1}{l}{SFT (CS) + DPO (Turn)} & 38.68 {\scriptsize \textcolor{green!60!black}{\rule{1.44pt}{5pt}}} & 67.59 {\scriptsize \textcolor{green!60!black}{\rule{3.02pt}{5pt}}} & 57.20 {\scriptsize \textcolor{red}{\rule{0.42pt}{5pt}}} & 50.65 {\scriptsize \textcolor{green!60!black}{\rule{3.28pt}{5pt}}} & 86.29 {\scriptsize \textcolor{green!60!black}{\rule{4.00pt}{5pt}}} & 58.70 {\scriptsize \textcolor{green!60!black}{\rule{1.13pt}{5pt}}} \\
\hline
\textbf{Qwen-2-7B} &&&&&& \\
Baseline & 8.14 & 32.62 & 37.91 & 25.15 & 43.57 & 36.94 \\
SFT (CS) & 32.99 {\scriptsize \textcolor{green!60!black}{\rule{24.85pt}{5pt}}} & 58.56 {\scriptsize \textcolor{green!60!black}{\rule{25.94pt}{5pt}}} & 43.12 {\scriptsize \textcolor{green!60!black}{\rule{5.21pt}{5pt}}} & 19.99 {\scriptsize \textcolor{red}{\rule{5.16pt}{5pt}}} & 42.21 {\scriptsize \textcolor{red}{\rule{1.36pt}{5pt}}} & 38.17 {\scriptsize \textcolor{green!60!black}{\rule{1.23pt}{5pt}}} \\
SFT (CS) + DPO (Dial.) & 10.82 {\scriptsize \textcolor{green!60!black}{\rule{2.68pt}{5pt}}} & 28.58 {\scriptsize \textcolor{red}{\rule{4.04pt}{5pt}}} & 30.11 {\scriptsize \textcolor{red}{\rule{7.80pt}{5pt}}} & 13.20 {\scriptsize \textcolor{red}{\rule{11.95pt}{5pt}}} & 24.08 {\scriptsize \textcolor{red}{\rule{19.49pt}{5pt}}} & 39.69 {\scriptsize \textcolor{green!60!black}{\rule{2.75pt}{5pt}}} \\
SFT (CS) + DPO (Turn.) & 22.31 {\scriptsize \textcolor{green!60!black}{\rule{14.17pt}{5pt}}} & 45.21 {\scriptsize \textcolor{green!60!black}{\rule{12.59pt}{5pt}}} & 39.09 {\scriptsize \textcolor{green!60!black}{\rule{1.18pt}{5pt}}} & 21.81 {\scriptsize \textcolor{red}{\rule{3.34pt}{5pt}}} & 40.65 {\scriptsize \textcolor{red}{\rule{2.92pt}{5pt}}} & 32.89 {\scriptsize \textcolor{red}{\rule{4.05pt}{5pt}}} \\

\bottomrule
\end{tabular}
}
\caption{\textbf{Gameplay results.} Clemscore, average percentage of completed games, and average quality score.}
\label{tab:games-results-arxiv}
\vspace{1em}
\end{table*}

\begin{table*}[]
\centering
\resizebox{\textwidth}{!}{%
\begin{tabular}{llllllll}
\toprule
 & \multicolumn{1}{c}{\textbf{In Domain}} & \multicolumn{1}{c}{\textbf{Out of Domain}} & \multicolumn{5}{c}{\textbf{Functional, Formal, General, Instruction Following}} \\ \hline
\multicolumn{1}{l}{\textbf{Model}} & \textbf{Clemscore} & \textbf{Clemscore} & \textbf{Executive} & \textbf{Socio-Emo} & \textbf{GLUE D.} & \textbf{General QA} & \textbf{IFEval} \\
\midrule
\textbf{Llama-3.1-8B} &&&&&& \\
\multicolumn{1}{l}{Baseline} & 19.39 & 24.58 & 39.24 & 57.16 & 38.06 & 41.86 & 76.88 \\
\multicolumn{1}{l}{SFT (CS)} & 40.11 {\scriptsize \textcolor{green!60!black}{\rule{20.72pt}{5pt}}} & 22.53 {\scriptsize \textcolor{red}{\rule{2.05pt}{5pt}}} & 39.93 {\scriptsize \textcolor{green!60!black}{\rule{0.69pt}{5pt}}} & 59.51 {\scriptsize \textcolor{green!60!black}{\rule{2.35pt}{5pt}}} & 40.43 {\scriptsize \textcolor{green!60!black}{\rule{2.37pt}{5pt}}} & 29.95 {\scriptsize \textcolor{red}{\rule{11.91pt}{5pt}}} & 67.25 {\scriptsize \textcolor{red}{\rule{9.63pt}{5pt}}} \\
\multicolumn{1}{l}{SFT (CS) + DPO (Dial.)} & 32.33 {\scriptsize \textcolor{green!60!black}{\rule{12.94pt}{5pt}}} & 19.50 {\scriptsize \textcolor{red}{\rule{5.08pt}{5pt}}} & 38.50 {\scriptsize \textcolor{red}{\rule{0.74pt}{5pt}}} & 55.10 {\scriptsize \textcolor{red}{\rule{2.06pt}{5pt}}} & 36.20 {\scriptsize \textcolor{red}{\rule{1.86pt}{5pt}}} & 26.57 {\scriptsize \textcolor{red}{\rule{15.29pt}{5pt}}} & 68.39 {\scriptsize \textcolor{red}{\rule{8.49pt}{5pt}}} \\

SFT (CS) + DPO (Turn.) & 39.65 {\scriptsize \textcolor{green!60!black}{\rule{20.26pt}{5pt}}} & 23.81 {\scriptsize \textcolor{red}{\rule{0.77pt}{5pt}}} & 39.90 {\scriptsize \textcolor{green!60!black}{\rule{0.66pt}{5pt}}}   & 59.75 {\scriptsize \textcolor{green!60!black}{\rule{2.59pt}{5pt}}}  & 37.83 {\scriptsize \textcolor{red}{\rule{0.23pt}{5pt}}}  & 31.61 {\scriptsize \textcolor{red}{\rule{10.25pt}{5pt}}}  & 70.00 {\scriptsize \textcolor{red}{\rule{6.88pt}{5pt}}}  \\

\multicolumn{1}{l}{GRPO} & 24.89 {\scriptsize \textcolor{green!60!black}{\rule{5.50pt}{5pt}}} & 33.92 {\scriptsize \textcolor{green!60!black}{\rule{9.34pt}{5pt}}} & 39.39 {\scriptsize \textcolor{green!60!black}{\rule{0.15pt}{5pt}}} & 57.51 {\scriptsize \textcolor{green!60!black}{\rule{0.35pt}{5pt}}} & 38.68 {\scriptsize \textcolor{green!60!black}{\rule{0.62pt}{5pt}}} & 41.52 {\scriptsize \textcolor{red}{\rule{0.34pt}{5pt}}} & 76.67 {\scriptsize \textcolor{red}{\rule{0.21pt}{5pt}}} \\
\multicolumn{1}{l}{SFT (CS) + GRPO} & 24.30 {\scriptsize \textcolor{green!60!black}{\rule{4.91pt}{5pt}}} & 31.81 {\scriptsize \textcolor{green!60!black}{\rule{7.23pt}{5pt}}} & 33.35 {\scriptsize \textcolor{red}{\rule{5.89pt}{5pt}}} & 58.67 {\scriptsize \textcolor{green!60!black}{\rule{1.51pt}{5pt}}} & 37.31 {\scriptsize \textcolor{red}{\rule{0.75pt}{5pt}}} & 42.82 {\scriptsize \textcolor{green!60!black}{\rule{0.96pt}{5pt}}} & 75.77 {\scriptsize \textcolor{red}{\rule{1.11pt}{5pt}}} \\ 
\\
\textbf{Llama-3.1-70B} &&&&&& \\
\multicolumn{1}{l}{Baseline} &  37.24 & 47.37  & 52.42 & 71.37 & 46.16 & 60.56 & 85.16 \\
\multicolumn{1}{l}{SFT (CS)} & 53.60 {\scriptsize \textcolor{green!60!black}{\rule{16.37pt}{5pt}}} & 54.40 {\scriptsize \textcolor{green!60!black}{\rule{7.03pt}{5pt}}}  & 55.17 {\scriptsize \textcolor{green!60!black}{\rule{2.75pt}{5pt}}} & 69.25 {\scriptsize \textcolor{red}{\rule{2.25pt}{5pt}}} & 47.72 {\scriptsize \textcolor{green!60!black}{\rule{1.57pt}{5pt}}} & 44.91 {\scriptsize \textcolor{red}{\rule{15.65pt}{5pt}}} & 79.38 {\scriptsize \textcolor{red}{\rule{5.78pt}{5pt}}} \\
\multicolumn{1}{l}{SFT (CS) + DPO (Dial.)} & 36.92 {\scriptsize \textcolor{red}{\rule{0.32pt}{5pt}}} & 45.46 {\scriptsize \textcolor{red}{\rule{1.91pt}{5pt}}} & 48.94 {\scriptsize \textcolor{red}{\rule{3.48pt}{5pt}}}  & 67.89 {\scriptsize \textcolor{red}{\rule{3.48pt}{5pt}}} & 37.73 {\scriptsize \textcolor{red}{\rule{8.43pt}{5pt}}} & 38.78 {\scriptsize \textcolor{red}{\rule{21.78pt}{5pt}}} & 82.26 {\scriptsize \textcolor{red}{\rule{2.90pt}{5pt}}}  \\
\multicolumn{1}{l}{SFT (CS) + DPO (Turn)} & 38.68 {\scriptsize \textcolor{green!60!black}{\rule{1.44pt}{5pt}}} & 50.65 {\scriptsize \textcolor{green!60!black}{\rule{3.28pt}{5pt}}} & 50.22 {\scriptsize \textcolor{red}{\rule{2.20pt}{5pt}}} & 70.21 {\scriptsize \textcolor{red}{\rule{1.16pt}{5pt}}} & 39.23 {\scriptsize \textcolor{red}{\rule{6.93pt}{5pt}}} & 44.86 {\scriptsize \textcolor{red}{\rule{15.70pt}{5pt}}} & 85.68 {\scriptsize \textcolor{green!60!black}{\rule{0.52pt}{5pt}}}\\
\\
\textbf{Qwen-2-7B} &&&&&& \\
Baseline & 08.14 & 25.04 & 40.77 & 56.58 & 59.68 & 16.83 & 59.16 \\
SFT (CS) & 32.99 {\scriptsize \textcolor{green!60!black}{\rule{24.85pt}{5pt}}} & 19.99 {\scriptsize \textcolor{red}{\rule{5.16pt}{5pt}}} & 40.42 {\scriptsize \textcolor{red}{\rule{0.35pt}{5pt}}} & 56.16 {\scriptsize \textcolor{red}{\rule{0.42pt}{5pt}}} & 52.98 {\scriptsize \textcolor{red}{\rule{6.7pt}{5pt}}} & 21.36 {\scriptsize \textcolor{green!60!black}{\rule{4.53pt}{5pt}}} & 53.87 {\scriptsize \textcolor{red}{\rule{5.29pt}{5pt}}} \\
SFT (CS) + DPO (Dial.) & 10.82 {\scriptsize \textcolor{green!60!black}{\rule{2.68pt}{5pt}}} & 13.20 {\scriptsize \textcolor{red}{\rule{11.95pt}{5pt}}} & 39.31 \textcolor{red}{\rule{1.46pt}{5pt}} & 56.79 {\scriptsize \textcolor{green!60!black}{\rule{0.215pt}{5pt}}} & 53.74 {\scriptsize \textcolor{red}{\rule{5.94pt}{5pt}}} & 15.35 {\scriptsize \textcolor{red}{\rule{1.48pt}{5pt}}} & 54.20 {\scriptsize \textcolor{red}{\rule{4.96pt}{5pt}}} \\
SFT (CS) + DPO (Turn.) & 22.31 {\scriptsize \textcolor{green!60!black}{\rule{14.17pt}{5pt}}} & 21.81 {\scriptsize \textcolor{red}{\rule{3.34pt}{5pt}}} & 39.22 \textcolor{red}{\rule{1.55pt}{5pt}} & 58.83 {\scriptsize \textcolor{green!60!black}{\rule{0.25pt}{5pt}}} & 48.34 {\scriptsize \textcolor{red}{\rule{11.34pt}{5pt}}} & 13.71 {\scriptsize \textcolor{red}{\rule{3.12pt}{5pt}}} & 54.99 {\scriptsize \textcolor{red}{\rule{4.17pt}{5pt}}} \\
\end{tabular}
}
\caption{\textbf{Main results}. Clemscore, average percentage of completed games, and average quality score. We report the best SFT variant Cold Start (CS). Executive includes LogiQA 2.0, CLadder, and WinoGrande. Socio-emotional includes EQ-Bench, LM-Pragmatics, SocialIQA, and SimpleToM.  General QA includes MMLU-Pro and BBH, while  IFEval targets instruction-following specifically. Formal capabilities are evaluated in GLUE Diagnostics. Colored bars indicate whether there is a positive (\textit{green}) or negative difference (\textit{red}) wrt. the Baseline model.}
\label{tab:results-overview}
\end{table*}

\subsection{Interactive Gameplay}

Table \ref{tab:games-results-arxiv} shows dialogue game performance across all evaluated models. 
The Llama-3.1-8B \texttt{Baseline} demonstrates basic rule-following, completing just over half of the games in both in-domain and out-of-domain settings; however, its gameplay quality is generally low.
Interestingly, the \texttt{Baseline} achieves higher quality and \texttt{clemscore} on out-of-domain games, suggesting that the in-domain set may be inherently more challenging for this model.
Training with SFT on successful episodes improves in-domain performance but reduces generalisation, with a decrease in all dimensions of performance on out-of-domain games.
This aligns with prior observations that SFT tends to overfit to the training distribution \citep{zeng2023agenttuning, chu2025sft, setlur_scaling_2025}.
Turning to DPO, we observe that both its variants (with dialogue- or turn-level preference pairs) improve in-domain performance over the \texttt{Baseline} but fail to outperform the best SFT model. Moreover, they suffer from even stronger degradation in out-of-domain games. We believe this could be a result of ``likelihood displacement"---a weakness of DPO-based training strategies~\citep{razin2024unintentional}.
We applied SFT and DPO to Qwen-2-7B as well. First of all, we notice that the baseline version is less performant than Llama. Despite this difference, we observe a similar trend with respect to the post-training methods, with SFT and DPO showing an improvement over the baseline for in-domain games, but not on out-of-domain games. 
The most robust training regime is GRPO. 
Applied directly on the base model, it leads to consistent improvements ($+5.50$ in in-domain \texttt{clemscore} and $+9.34$ out-of-domain), albeit with a slight decrease in the number of in-domain games played. 
When applied on top of SFT, GRPO recovers from this slight drop, likely leveraging the SFT model's stronger ability to adhere to game instructions, though at the cost of more modest gains in quality score.

\paragraph{What happens at the 70B scale?} 
Larger models are known to possess stronger instruction-following capabilities, a skill that is especially relevant for our benchmark, where accurate interpretation of game prompts is critical to gameplay.
We therefore conduct additional experiments with \texttt{Llama-3.1-70B}, applying the same training regimes used for the 8B model, but excluding GRPO due to its high computational cost and our resource constraints.
It is worth noting that simply using this model as a starting point nearly doubles the overall \texttt{clemscore} on both in-domain and out-of-domain games (results are shown in the bottom half of Table~\ref{tab:games-results-arxiv}). 
When applying SFT to the larger base model, we observe diminishing returns for in-domain games, with an improvement of $16.36$ points over the \texttt{Baseline}---lower than the increase obtained when applying SFT on the base 8B model ($20.72$). However, on out-of-domain games, we record our best-scoring model with a \texttt{clemscore} of $54.4$. To calibrate this result, this is still far below the top leaderboard \texttt{clemscore} of $70$, achieved by \texttt{o3-mini-2025-01-31}.\footnote{Based on \url{https://clembench.github.io/leaderboard.html}, accessed 2025-05-14.}
Finally, we find that applying DPO on top of SFT reverses some of the gains of SFT alone. Between the two DPO variants, DPO with dialogue-level preference data obtains lower scores on both in-domain and out-of-domain games---another possible case of overfitting after the preliminary SFT phase. 
The turn-level variant of DPO yields modest improvements over the \texttt{Baseline} on both in-domain and out-of-domain games, but it still falls short of the best SFT model.

\subsection{Non-Interactive Benchmarks}

While our main focus is on performance in dialogue games, we also evaluate models across a broad set of other tasks.
This helps identify whether training on dialogue games leads to regressions in general language skills (e.g., formal competence) or, conversely, contributes to improvements in language use (e.g., functional competence).
Table~\ref{tab:results-overview} summarises the results across these evaluations.

Among the post-training regimes tested, GRPO is the most balanced overall, with lower oscillations---either improvements or regressions---in non-interactive task performance. Tables~\ref{tab:reference-based-results-detailed-1} and \ref{tab:reference-based-results-detailed-2} in Appendix~\ref{sec:appendix-eval-instruction-following} give a complete overview of the results. Here, we highlight that training on dialogue games seems to provide a modest improvement on the ``Executive'' task category for the 70B model trained with \texttt{SFT (CS)}, 
suggesting that learning from dialogue games may enhance a model's ability to integrate and reason over contextual information---an ability \citet{momente2025triangulating} identified as critical for these tasks. 

Another relevant finding from this evaluation is the substantial drop in the ability to follow instructions, as measured by IFEval. 
This calls for further investigation into instruction-following training regimes that are more robust to interactive settings, allowing models not only to handle single-turn prompts---as is common in current instruction-following regimes---but also to participate effectively in complex, goal-oriented, rule-governed, and multi-turn tasks such as dialogue games. 

\subsection{Qualitative Discussion}

An outcome of our evaluation is that increasing the size of the LLM backbone does correspond to better instruction-following abilities in both traditional benchmarks and on our dialogue games. However, current models still do not show desirable online adaptation skills (also see Appendix~\ref{app:error_analysis} for a detailed error analysis on gameplay abilities). 
Thanks to language prompts describing a game $g$, we should be able to derive a game policy $\pi_g$ on the fly. In some cases, this might not be enough, and therefore, it is possible to use SFT to learn how to play the game by mimicking transcripts. However, because the model does not have the chance to play by itself, it might miss some nuances of the game and overfit on specific rules/formats of the game at hand. On the other hand, thanks to the online training regime of GRPO, it is possible to acquire general-purpose instruction following abilities that allow models to perform better in out-of-domain games as well as retain abilities required for more general-purpose NLP tasks---a result in line with test-time compute analysis for RL algorithms reported in the literature \citep{chu2025sft, setlur_scaling_2025}. We report additional dialogue transcripts in Table~\ref{app:dialogue_games_transcripts} in the Appendix.

\section{Conclusion}

In this paper, we explore to what extent synthetic interaction in what we call Dialogue Games---goal-directed and rule-governed activities driven by verbal actions---can provide a learning signal for LLMs. 
We created \textsc{playpen}, an environment that facilitates synthetic data generation of dialogue transcripts that can be used to train LLMs. 
We provide an extensive evaluation of a variety of state-of-the-art post-training methods such as SFT, DPO, and GRPO, and show how GRPO is a more stable training regime that prevents overfitting to in-domain games and facilitates generalisation to out-of-domain dialogue games. Additionally, we demonstrate that when leveraging dialogue games, it is possible to observe a performance improvement when completing more traditional, non-interactive general instruction-following tasks such as MMLU-Pro.

The framework and the baselines presented here can form the basis for exciting future work, for example investigating novel training regimes based on reinforcement learning to truly leverage the multi-turn nature of dialogue games, or exploring the use of intermediate language feedback that can be acquired as part of the interaction (along the lines of \citealt{sumers2021learning}), and further exploring the potential of the ``learning in conversational interaction'' paradigm.


\section*{Limitations}

Our study makes significant strides in demonstrating the potential of dialogue games as a valuable source of feedback signals for training LLMs. The \textsc{playpen} environment offers a versatile platform for exploring both off- and online learning paradigms, and our comparative analysis of post-training methods, including SFT, DPO, and reinforcement learning with GRPO, provides a strong foundation for future research. 

However, the current study has several limitations that warrant further investigation. Firstly, in our Direct Preference Optimization (DPO) experiments, we utilize a seed dataset of successful dialogues from which we derive positive and negative pairs. However, for the turn variant, we assume that all turns within these dialogues are successful. This assumption may not hold true in real-world scenarios, particularly when corrections or clarifications are present within the dialogue~\citep{chiyah2024repairs}. Secondly, our work does not explore multi-turn training methods, which could be crucial for more complex dialogue games and real-world applications where it is important to perform credit-assignment across multiple turns \citep[e.g.,][]{zhou_archer_2024}.

Additionally, we did not incorporate reasoning models ~\citep[e.g.,][]{deepseekai2025deepseekr1incentivizingreasoningcapability} or chain-of-thought prompting techniques~\citep{wei2022chain}, which have shown promise in enhancing LLM performance in other tasks. Furthermore, our evaluation of GRPO is limited to a smaller 8B LLM. Evaluating the effectiveness of GRPO on larger models, such as the 70B parameter model, would provide valuable insights into the scalability of our findings. Unfortunately, due to limited computational resources, we leave this exploration for future work.

The current set of dialogue games in \textsc{Playpen} provides a foundation for our research, but it is not exhaustive. Future work should aim to expand the set of games to be more representative of the diverse range of language games encountered in real-world scenarios. This is especially important considering that \citet{momente2025triangulating} has demonstrated that dialogue games are actually more discriminative than other benchmarks because they likely require important underlying capabilities such as working memory.

Finally, our focus is on dialogue game feedback, which is inherent to the task itself. We do not consider additional forms of feedback, such as explicit corrective feedback, which could potentially enhance learning, as explored in prior work \citep[e.g.,][]{sumers2021learning, mccallum2023feedback, xi2024teaching}.

\subsection*{Ethical considerations}

Our work broadly falls under the rubric of ``self-improvement" of language models. There is a small, but non-zero chance that such self-improvement, if run unsupervised and in recursive loops, might lead to uncontrolled gains. Our advice hence would be to define clear stopping criteria for learning runs. Additionally, we created \textsc{Playpen} as a synthetic and simulated learning environment where the model doesn't have access to external tools or, more broadly, it doesn't have the ability to execute actions in the real world.

 \section*{Acknowledgments}
This research has been partially supported and sponsored by the Province of Bolzano and the EU through the project Artificial Intelligence Laboratory (AI-Lab)
ERDF 2021-2027 EFRE1047, CUP: I53C23001670009 (01/01/2024 - 31/12/2026). Alessandro Suglia, Raffaella Bernardi, and David Schlangen acknowledge ISCRA for awarding this project access to the LEONARDO supercomputer, owned by the EuroHPC Joint Undertaking, hosted by CINECA (Italy). 
Mario Giulianelli was in part supported by an ETH Zurich Postdoctoral Fellowship. Alberto Testoni is funded by the project CaRe-NLP with file number NGF.1607.22.014 of the research programme AiNed Fellowship Grants, which is (partly) financed by the Dutch Research Council (NWO). Raquel Fern\'andez was supported by the European Research Council (ERC Consolidator Grant DREAM 819455). David Schlangen acknowledges support by Deutsche Forschungsgemeinschaft (DFG, German Research Foundation) – grant 423217434 (RECOLAGE), and University of Potsdam (President's Fund).\looseness-1 


\bibliography{custom}

\appendix
\section*{Appendix} 

In the following, we report additional details for our paper, starting from the dialogue games used in our work (see Section \ref{app:games} for all the available games and Section \ref{app:dialogue_games_transcripts} for some examples). We also provide details of our experimental protocol, including training data (see Section \ref{app:training_data_generation}), our training regime (see Section \ref{app:training_details}), and additional results (see Section \ref{app:additional_results}). Beyond raw scores on the different benchmarks that are part of Playpen, we also provide a comprehensive error analysis to understand the behaviour of each model after our post-training regime via Playpen (see Section \ref{app:error_analysis}). Finally, we also report licenses for our code and data (see Section \ref{app:license}).

\section{Dialogue Games}
\label{app:games}
\begin{itemize}
    \item \textbf{Taboo}: two-player game where one player gives a clue, not using certain words, and the other player needs to guess a target word based on the clue. 
    
    \item \textbf{PrivateShared}: two-player scorekeeping game where an answerer agent goes through a form with a questioner. The GameMaster keeps track of which information has been already shared and checks whether players also do so correctly.  
    
    \item \textbf{ImageGame}: two-player instruction giving and following game where one player provides a description of a target image (a matrix in ASCII format) and the other player has to draw (i.e. generate some matrix again in ASCII format) an image based on the description. The generated image should match the target one.
    
    \item \textbf{ReferenceGame}: two-player game where one player is given three different images (one of them is selected as the target) and has to generate a referring expression that describes the target image by differentiating it from other two (distractors). Another player is then given the same images (orders are shuffled) and is asked to guess which one is the target based on the given referring expression. It is the only single-turn game in the benchmark.
    
    \item \textbf{Wordle}: popular single-player game where the task is to guess a 5-letter word. In each turn, feedback is provided based on the placements of characters in the guessing attempts.
    
    \item \textbf{Wordle With Clue}: slightly changed version of the base Wordle game with the addition of a clue for the target word.
    \item \textbf{Wordle With Critic}: a two-player version of the base Wordle game where the second player (critic) provides feedback on the guesses of the first player.

    \item \textbf{Codenames}: a popular cooperative game with two teams that try to uncover their agent's code names (words). Here, one player has to describes clues 
    that could strategically correspond to more than one word. Teams are composed of a Spymaster that provides clues and a Field operative tasked with guessing. In our case, two LLMs are placed on the same team, with the other team being characterized by programmatic behaviour.
    
    \item \textbf{AdventureGame}: single-player text adventure game where a player is placed in a random location of an environment and is given a task (e.g. pick up the flower and place it on the table). The player explores the environment by giving commands (e.g. ``go to the kitchen'', ``open the cupboard'') and the environment provides feedback regarding whether the command can be realised and its outcome or not. The player explores the world (with multiple rooms and objects in them) and has to decide on its own when to stop the game.

    \item \textbf{GuessWhat}: two-player information seeking game where one player needs to guess the target word (out of eight options) by asking questions about the target. The other player knows the target word and answers the questions with ``yes'' or ``no''.
    
    \item \textbf{Matchit}: two-player game where each player is given an image (an ASCII representation, e.g. grid), which is not revealed to the other player. The goal is for players to understand whether they are looking at the same image or not. Players are allowed to ask questions to each other about the image and provide answers to the questions.
    
    \item \textbf{Map Navigation}: single-player game where a player has to navigate a map by indicating the directions where to go. Initially, the player is placed in a random room and is given information regarding where adjacent rooms are located (east, west, north, south) and about the name of the current room (e.g kitchen, bathroom, garage). Every time the player makes a step, is provided with these information updated at the current location. When the player thinks it has explored everything in the map, it should decided to stop the navigation on its own.

    \item \textbf{Map Navigation - specific room}: modified version of the game above. The player is given the name of a target room (e.g. kitchen, garage, living room) and is required to stop exploring once it finds the target.

    \item \textbf{Map Navigation} with graph reasoning: modified version of the base Mapnavigation game where, along with the already described exploration mechanism, at each step the player is required to draw (i.e. generate JSON-like structure) the graph representing the map discovered up to that point.
\end{itemize}

\section{License for Artifacts}
\label{app:license}

\textsc{Playpen}'s implementation reuses important components from clemcore, and all the games used come from clembench~\citep{Chalamalasetti-2023}. Therefore, we inherit its MIT License for \textsc{Playpen}. We acknowledge that the transcripts used for SFT and DPO training are derived from the clembench-runs~\footnote{\url{https://github.com/clembench/clembench-runs/blob/main/LICENSE}} repo which has CC-By 4.0 License.

\section{Dialogue Games Transcripts} 
\label{app:dialogue_games_transcripts}
We include below sample transcripts for three of the games present in this work: GuessWhat, Taboo, and Codenames. The examples show successful episodes for the Taboo and GuessWhat games, while for Codenames the episode is aborted before reaching the end.

\newcounter{utterance}


\subsection{GuessWhat game Transcript}

{ \footnotesize  \setcounter{utterance}{1}
\setlength{\tabcolsep}{0pt}
\begin{supertabular}{c@{$\;$}|p{.05\linewidth}@{}p{.15\linewidth}p{.15\linewidth}p{.15\linewidth}p{.15\linewidth}p{.15\linewidth}}

    \# & $\;$A & \multicolumn{4}{c}{Game Master} & $\;\:$B\\
    \hline 

    \theutterance \stepcounter{utterance}  

    & & \multicolumn{4}{p{0.95\linewidth}}{\cellcolor[rgb]{0.9,0.9,0.9}{%
	\makecell[{{p{\linewidth}}}]{%
	  \tt {\tiny [A$\langle$GM]}  
	 Let's play a guessing game. Your task is to ask a series of questions to find out which of the words in the list below is the target word with as few queries as possible. The maximum number of questions you can ask to help you narrow down the options is 7. Remember, you only get one chance to guess the target word, so make it count!\\ \tt \\ \tt The rules of the game are: \\ \tt \\ \tt 1. Only ask one question per turn. DO NOT ASK MORE THAN ONE QUESTION.\\ \tt \\ \tt 2. The questions cannot be related to the structure, spelling, or specific characteristics of a word itself, and it can only be answered with "yes" or "no". \\ \tt \\ \tt 3. Use the following format for asking questions and DO NOT ADD ANY TEXT OTHER THAN THIS: \\ \tt \\ \tt QUESTION: <your question>\\ \tt \\ \tt 4. You have only one opportunity to guess the target word. To make your guess, use the following format and DO NOT ADD ANY TEXT OTHER THAN THIS: \\ \tt \\ \tt GUESS: <target word>\\ \tt \\ \tt Remember you cannot ask for a hint. Make your question brief. Let’s start!\\ \tt \\ \tt List of candidate words: ['Mackerel', 'Grouper', 'Carp', 'Catfish', 'Halibut', 'Flounder', 'Minnow', 'Goldfish'] 
	  } 
	   } 
	   } 
	 & \\

    \theutterance \stepcounter{utterance}  

    & \multicolumn{4}{p{0.6\linewidth}}{\cellcolor[rgb]{0.8,1,0.9}{%
	\makecell[{{p{\linewidth}}}]{%
	  \tt {\tiny [A$\rangle$GM]}  
	 QUESTION: Is the target word a freshwater fish? 
	  } 
	   } 
	   } 
	 & & \\

    \theutterance \stepcounter{utterance}  

    & & \multicolumn{4}{p{0.8\linewidth}}{\cellcolor[rgb]{0.9,0.9,0.9}{%
	\makecell[{{p{\linewidth}}}]{%
	  \tt {\tiny [GM$\rangle$B]}  
	 Let’s play a guessing game! Your task is to answer the other player's questions. Based on your knowledge of the word: Catfish, respond to the following questions or guesses. Limit your response to only 'yes' or 'no' with no explanation or other words. Never reveal the answer in your response.\\ \tt \\ \tt You must reply using the format below and DO NOT ADD ANY TEXT OTHER THAN THIS:\\ \tt \\ \tt ANSWER: <some text>\\ \tt \\ \tt Target Word: Catfish\\ \tt \\ \tt QUESTION: Is the target word a freshwater fish? 
	  } 
	   } 
	   } 
	 & \\

    \theutterance \stepcounter{utterance}  

    & & & \multicolumn{4}{p{0.6\linewidth}}{\cellcolor[rgb]{1,0.85,0.72}{%
	\makecell[{{p{\linewidth}}}]{%
	  \tt {\tiny [GM$\langle$B]}  
	 ANSWER: Yes 
	  } 
	   } 
	   } 
	  \\



    \theutterance \stepcounter{utterance}  

    & \multicolumn{4}{p{0.6\linewidth}}{\cellcolor[rgb]{0.8,1,0.9}{%
	\makecell[{{p{\linewidth}}}]{%
	  \tt {\tiny [A$\rangle$GM]}  
	 QUESTION: Is the target word commonly kept as a pet? 
	  } 
	   } 
	   } 
	 & & \\



    \theutterance \stepcounter{utterance}  

    & & & \multicolumn{4}{p{0.6\linewidth}}{\cellcolor[rgb]{1,0.85,0.72}{%
	\makecell[{{p{\linewidth}}}]{%
	  \tt {\tiny [GM$\langle$B]}  
	 ANSWER: Yes 
	  } 
	   } 
	   } 
	  \\



    \theutterance \stepcounter{utterance}  

    & \multicolumn{4}{p{0.6\linewidth}}{\cellcolor[rgb]{0.8,1,0.9}{%
	\makecell[{{p{\linewidth}}}]{%
	  \tt {\tiny [A$\rangle$GM]}  
	 QUESTION: Is the target word relatively small in size? 
	  } 
	   } 
	   } 
	 & & \\



    \theutterance \stepcounter{utterance}  

    & & & \multicolumn{4}{p{0.6\linewidth}}{\cellcolor[rgb]{1,0.85,0.72}{%
	\makecell[{{p{\linewidth}}}]{%
	  \tt {\tiny [GM$\langle$B]}  
	 ANSWER: No 
	  } 
	   } 
	   } 
	  \\



    \theutterance \stepcounter{utterance}  

    & \multicolumn{4}{p{0.6\linewidth}}{\cellcolor[rgb]{0.8,1,0.9}{%
	\makecell[{{p{\linewidth}}}]{%
	  \tt {\tiny [A$\rangle$GM]}  
	 QUESTION: Is the target word native to North America? 
	  } 
	   } 
	   } 
	 & & \\



    \theutterance \stepcounter{utterance}  

    & & & \multicolumn{4}{p{0.6\linewidth}}{\cellcolor[rgb]{1,0.85,0.72}{%
	\makecell[{{p{\linewidth}}}]{%
	  \tt {\tiny [GM$\langle$B]}  
	 ANSWER: Yes 
	  } 
	   } 
	   } 
	  \\



    \theutterance \stepcounter{utterance}  

    & \multicolumn{4}{p{0.6\linewidth}}{\cellcolor[rgb]{0.8,1,0.9}{%
	\makecell[{{p{\linewidth}}}]{%
	  \tt {\tiny [A$\rangle$GM]}  
	 QUESTION: Is the target word a scaleless fish? 
	  } 
	   } 
	   } 
	 & & \\



    \theutterance \stepcounter{utterance}  

    & & & \multicolumn{4}{p{0.6\linewidth}}{\cellcolor[rgb]{1,0.85,0.72}{%
	\makecell[{{p{\linewidth}}}]{%
	  \tt {\tiny [GM$\langle$B]}  
	 ANSWER: Yes 
	  } 
	   } 
	   } 
	  \\



    \theutterance \stepcounter{utterance}  

    & \multicolumn{4}{p{0.6\linewidth}}{\cellcolor[rgb]{0.8,1,0.9}{%
	\makecell[{{p{\linewidth}}}]{%
	  \tt {\tiny [A$\rangle$GM]}  
	 QUESTION: Does the target word have whisker-like barbels? 
	  } 
	   } 
	   } 
	 & & \\



    \theutterance \stepcounter{utterance}  

    & & & \multicolumn{4}{p{0.6\linewidth}}{\cellcolor[rgb]{1,0.85,0.72}{%
	\makecell[{{p{\linewidth}}}]{%
	  \tt {\tiny [GM$\langle$B]}  
	 ANSWER: Yes 
	  } 
	   } 
	   } 
	  \\



    \theutterance \stepcounter{utterance}  

    & \multicolumn{4}{p{0.6\linewidth}}{\cellcolor[rgb]{0.8,1,0.9}{%
	\makecell[{{p{\linewidth}}}]{%
	  \tt {\tiny [A$\rangle$GM]}  
	 GUESS: Catfish 
	  } 
	   } 
	   } 
	 & & \\

    \theutterance \stepcounter{utterance}  

    & & & \multicolumn{2}{p{0.5\linewidth}}{\cellcolor[rgb]{0.95,0.95,0.95}{%
	\makecell[{{p{\linewidth}}}]{%
	  \tt {\tiny [GM$|$GM]}  
	 CORRECT GUESS 
	  } 
	   } 
	   } 
	 & & \\

    \theutterance \stepcounter{utterance}  

    & & & \multicolumn{2}{p{0.5\linewidth}}{\cellcolor[rgb]{0.95,0.95,0.95}{%
	\makecell[{{p{\linewidth}}}]{%
	  \tt {\tiny [GM$|$GM]}  
	   End game.
	  } 
	   } 
	   } 
	 & & \\

\end{supertabular}
}


\subsection{Taboo game Transcript}

{ \footnotesize  \setcounter{utterance}{1}
\setlength{\tabcolsep}{0pt}
\begin{supertabular}{c@{$\;$}|p{.05\linewidth}@{}p{.15\linewidth}p{.15\linewidth}p{.15\linewidth}p{.15\linewidth}p{.15\linewidth}}

    \# & $\;$A & \multicolumn{4}{c}{Game Master} & $\;\:$B\\
    \hline 

    \theutterance \stepcounter{utterance}  

    & & \multicolumn{4}{p{0.90\linewidth}}{\cellcolor[rgb]{0.9,0.9,0.9}{%
	\makecell[{{p{\linewidth}}}]{%
	  \tt {\tiny [A$\langle$GM]}  
	 You are playing a collaborative word guessing game in which you have to describe a target word for another player to guess.\\ \tt \\ \tt Rules:\\ \tt (a) You have to reply in the form: CLUE: <some text>. Guesses from the other player will start with GUESS.\\ \tt (b) You cannot use the target word itself, parts or morphological variants of it in your description.\\ \tt (c) In addition, the same rules apply for related words which are provided below.\\ \tt \\ \tt End conditions:\\ \tt (i) If you use the target word or a related word in your description, then you lose.\\ \tt (ii) If the other player can guess the target word in 3 tries, you both win.\\ \tt \\ \tt Let us start. This is the target word that you need to describe and that the other player needs to guess:\\ \tt \\ \tt ordinary\\ \tt \\ \tt Related words are:\\ \tt \\ \tt - common\\ \tt - normal\\ \tt - plain\\ \tt \\ \tt Important: You are under time pressure, give short descriptions that are to the point! 
	  } 
	   } 
	   } 
	 & \\

    \theutterance \stepcounter{utterance}  

    & \multicolumn{4}{p{0.6\linewidth}}{\cellcolor[rgb]{0.8,1,0.9}{%
	\makecell[{{p{\linewidth}}}]{%
	  \tt {\tiny [A$\rangle$GM]}  
	 CLUE: Something that is usual or expected. 
	  } 
	   } 
	   } 
	 & & \\





    \theutterance \stepcounter{utterance}  

    & & \multicolumn{4}{p{0.9\linewidth}}{\cellcolor[rgb]{0.9,0.9,0.9}{%
	\makecell[{{p{\linewidth}}}]{%
	  \tt {\tiny [GM$\rangle$B]}  
	 You are playing a collaborative word guessing game in which you have to guess a target word that another player describes to you.\\ \tt \\ \tt You can make one guess at each trial. You win when you guess the target word. You lose when you cannot guess it in 3 tries.\\ \tt \\ \tt After each trial you will get a new hint from the other player which starts with CLUE.\\ \tt \\ \tt Make your guesses by just saying the word using the following form: GUESS: <a word>\\ \tt \\ \tt Let us start.\\ \tt \\ \tt \\ \tt CLUE: Something that is usual or expected. 
	  } 
	   } 
	   } 
	 & \\

    \theutterance \stepcounter{utterance}  

    & & & \multicolumn{4}{p{0.6\linewidth}}{\cellcolor[rgb]{1,0.85,0.72}{%
	\makecell[{{p{\linewidth}}}]{%
	  \tt {\tiny [GM$\langle$B]}  
	 GUESS: Norm 
	  } 
	   } 
	   } 
	  \\







    \theutterance \stepcounter{utterance}  

    & \multicolumn{4}{p{0.6\linewidth}}{\cellcolor[rgb]{0.8,1,0.9}{%
	\makecell[{{p{\linewidth}}}]{%
	  \tt {\tiny [A$\rangle$GM]}  
	 CLUE: Not fancy or special. 
	  } 
	   } 
	   } 
	 & & \\







    \theutterance \stepcounter{utterance}  

    & & & \multicolumn{4}{p{0.6\linewidth}}{\cellcolor[rgb]{1,0.85,0.72}{%
	\makecell[{{p{\linewidth}}}]{%
	  \tt {\tiny [GM$\langle$B]}  
	 GUESS: Ordinary 
	  } 
	   } 
	   } 
	  \\

    \theutterance \stepcounter{utterance}  

    & & & \multicolumn{2}{p{0.5\linewidth}}{\cellcolor[rgb]{0.95,0.95,0.95}{%
	\makecell[{{p{\linewidth}}}]{%
	  \tt {\tiny [GM$|$GM]}  
	 CORRECT GUESS 
	  } 
	   } 
	   } 
	 & & \\



    \theutterance \stepcounter{utterance}  

    & & & \multicolumn{2}{p{0.5\linewidth}}{\cellcolor[rgb]{0.95,0.95,0.95}{%
	\makecell[{{p{\linewidth}}}]{%
	  \tt {\tiny [GM$|$GM]}  
	 End game.
	  } 
	   } 
	   } 
	 & & \\

\end{supertabular}
}

\subsection{Codenames game Transcript}

{ \footnotesize  \setcounter{utterance}{1}
\setlength{\tabcolsep}{0pt}
\begin{supertabular}{c@{$\;$}|p{.05\linewidth}@{}p{.15\linewidth}p{.15\linewidth}p{.15\linewidth}p{.15\linewidth}p{.15\linewidth}}

    \# & $\;$A & \multicolumn{4}{c}{Game Master} & $\;\:$B\\
    \hline

    \theutterance \stepcounter{utterance}  

    & & \multicolumn{4}{p{0.75\linewidth}}{\cellcolor[rgb]{0.9,0.9,0.9}{%
	\makecell[{{p{\linewidth}}}]{%
	  \tt {\tiny [A$\langle$GM]}  
	 Provide one single-word clue that relates to some of your team's words from the list below. You can choose to how many and to which words your clue relates to.\\ \tt The goal is to have your partner guess all of your team's words based on the clues you give them, before the other team has uncovered all of theirs. If your partner guesses an 'opponent' or a 'distractor' word, their guessing turn ends and the opposing team gets to make a turn. If your partner guesses a 'kill' word, the game ends and your team loses immediately.\\ \tt The more words your clue relates to, the quicker you can win the game, but the harder the associations may be to guess for your partner, so choose your clue word wisely. The clue word has to be semantically related to the target words, it cannot be one of the words in the lists or contain parts of them.\\ \tt \\ \tt Always give your single-word clue and your comma-separated list of related target words in the following format and make your answers as short as possible, never include any other text than is required in this form:\\ \tt \\ \tt CLUE: <WORD>\\ \tt TARGETS: <TARGETS>\\ \tt \\ \tt Your team words are: hail, neck, mountain, poland, sister, couch, stool, captain, officer.\\ \tt Your opponent's team words are: hair, teeth, england, sergeant, table, fog, desert, rye.\\ \tt Distractor words are: daughter, recliner, private, son, cousin, thigh, ireland.\\ \tt The kill words are: liquor. 
	  } 
	   } 
	   } 
	 & \\

    \theutterance \stepcounter{utterance}  

    & \multicolumn{4}{p{0.6\linewidth}}{\cellcolor[rgb]{0.8,1,0.9}{%
	\makecell[{{p{\linewidth}}}]{%
	  \tt {\tiny [A$\rangle$GM]}  
	 CLUE: storm\\ \tt TARGETS: hail, mountain 
	  } 
	   } 
	   } 
	 & & \\

    \theutterance \stepcounter{utterance}  

    & & \multicolumn{4}{p{0.9\linewidth}}{\cellcolor[rgb]{0.9,0.9,0.9}{%
	\makecell[{{p{\linewidth}}}]{%
	  \tt {\tiny [GM$\rangle$B]}  
	 Provide a comma-separated list of up to 2 words from the following list that best relate or are most closely associated with the word '\textbf{storm}'. \\ \tt Always start your list of guess(es) with 'GUESS: ' and do not include any other text in your answer.\\ \tt \\ \tt sister, hair, hail, couch, table, desert, recliner, poland, daughter, thigh, teeth, cousin, stool, sergeant, private, england, son, fog, ireland, liquor, neck, officer, rye, mountain, captain 
	  } 
	   } 
	   } 
	 & \\

    \theutterance \stepcounter{utterance}  

    & & & \multicolumn{4}{p{0.6\linewidth}}{\cellcolor[rgb]{1,0.85,0.72}{%
	\makecell[{{p{\linewidth}}}]{%
	  \tt {\tiny [GM$\langle$B]}  
	 GUESS: hail, storm 
	  } 
	   } 
	   } 
	  \\



\theutterance \stepcounter{utterance}  

    & & & \multicolumn{2}{p{0.7\linewidth}}{\cellcolor[rgb]{0.95,0.95,0.95}{%
	\makecell[{{p{\linewidth}}}]{%
	  \tt {\tiny [GM$|$GM]}  
	 Guessed word \textbf{storm} is the same word as the provided clue word (storm), you should only select words from the provided list
	  } 
	   } 
	   } 
	 & & \\ 
     
\theutterance \stepcounter{utterance}  

    & & & \multicolumn{2}{p{0.5\linewidth}}{\cellcolor[rgb]{0.95,0.95,0.95}{%
	\makecell[{{p{\linewidth}}}]{%
	  \tt {\tiny [GM$|$GM]}  
	 Game aborted.
	  } 
	   } 
	   } 
	 & & \\

\end{supertabular}
}

\section{Training Data Generation}
\label{app:training_data_generation}

\subsection{Supervised Fine-tuning}
\paragraph{Data Collection}
We preprocessed the dialogue transcripts available in the clembench-runs repository\footnote{\url{https://github.com/clembench/clembench-runs}} and transformed them into a format that allowed further filtering. The clembench-runs repository is a collection of all generated interaction data produced by the models evaluated on the benchmark.
The extraction included the removal of all GameMaster-specific logging and parsing prompts that were not part of the conversation pieces directly prompted to the model.
Besides the conversational data, we also extracted additional metadata such as the information about the model, played game, benchmark-version, experiment, and episode number, as well as whether the episode was successfully played, lost or aborted.

\paragraph{Data Filtering}
For the supervised fine-tuning, only successful episodes were considered for training, with lost and aborted ones discarded from the data.

\paragraph{Data Transformation}
After filtering, we added some game-specific data transformations to mitigate changes between the different benchmark versions and to improve training performance. Most of the transformations were necessary due to changes in the prompts between the benchmark versions and changes in the parsing rules for model answers. All of the transformations and associated justification are listed below:

\textbf{ImageGame}
While clembench versions 0.9 and 1.0 allowed the player to add \textit{"what is your next instruction"} to its answers, the same behavior led to parsing errors in the version 1.6, which resulted in the abortion of all ImageGame episodes.
To address this problem, all model answers from player-1 were truncated to only contain the correct format required by the version 1.6. 

\begin{lstlisting}[basicstyle=\small\ttfamily, breaklines=true, breakatwhitespace=true, breakindent=0pt]
Before: Instruction: Put a B in the first column of all rows
what is your next instruction

After: Instruction: Put a B in the first column of all rows
\end{lstlisting}

\textbf{Wordle and its variants}
For the three Wordle variants, there were a few successfully played episodes that contained an \textit{"INVALID\_FORMAT"} token inside the prompts. These instances were removed since the model should not reproduce outputs with invalid formats.
We have also removed the parts of episodes where the player is asked to provide again an answer after the providing an unparsable one (Wordle-specific).

\textbf{ReferenceGame}
For ReferenceGame, the initial prompt was changed from the older clembench versions (0.9 and 1.0) to version 1.6. While the older versions contained multiple examples (few-shot prompting), in version 1.6 there are no examples available. These examples directly biased the model towards a specific gameplay strategy, and provide a description of the grids (see following snippet from the old version of the prompt).

\begin{lstlisting}[basicstyle=\small\ttfamily, breaklines=true, breakatwhitespace=true, breakindent=0pt]
Here is an example with grids.
The first grid is the target grid and the following two grids are distractors.

Target grid:
X X X X X
O O X O O
O O X O O
O O X O O
O O X O O
...

The referring expression for the given target grid is like so:
Expression: Filled as T.
\end{lstlisting}

Here, the model is directly instructed to describe the whole grid as a letter or shape. However, not all grids follow this pattern. In addition, considering that the game is a two-player reference game, player-2 has a 33\% chance of guessing correctly the correct grid. This resulted in a situation where about 53\% of the successful episodes, player-1 described the target grid as "Filled as T" while, except for the prompt example, there is no T-shaped grid in the data. This meant that reference game data from the old benchmark versions could not be used for the training process due to the low quality. To mitigate this problem, data from the version 1.6 was used while 20-30\% of the episodes of each experiment were held out for testing.

\textbf{PrivateShared}
For PrivateShared, after the first experiments it appeared that in most cases, the trained model answered with \textit{"ASIDE: No"} to all probe-questions. In PrivateShared, the model should act like a customer at a travel agency that wants to travel. The agent asks questions about destination, time and other related properties of the inquiry. Turn after turn, the model has to tell the agent all the information the agent needs. After every question there is a block of probing questions where the model is asked whether or not specific information has been shared already, and the model has to answer with \textit{"ASIDE: yes"} or \textit{"ASIDE: no"}.

Considering the structure of the game, the model has to answer with \textit{"ASIDE: no"} to all probe-questions in the beginning of the game dialogue which changes to more and more \textit{"ASIDE: yes"} during the course of the game play depending on what information has already been shared. To prevent over-fitting, we have decided to reduce the number of probe-question answer pairs shown to the model by merging them into a single sample per iteration done in the main conversation.

The specific changes made to PrivateShared, ImageGame and ReferenceGame were partially derived by experimenting with them. For the rest of the games, no particular changes have been made.

\paragraph{Iterative Data-Processing}
While some of the previously described data transformations were motivated by observations made during the data preparation and collection phase, further experiments have been conducted to iteratively improve the data to optimize fine-tuning performance. This resulted in a final dataset which combined all the possible improvements learned through this process.

In total, more than 30 different experiments were conducted with different dataset configurations. They have been structured into nine main experiments, composed of one or more sub-experiments. Below is a description of the main experiments:

\textbf{D1} Contains, as an initial experiment, all successfully played episodes of all models. The dialogues are not processed in any way and just parsed into the model-specific chat-templates.

\textbf{D2} Contains only successfully played episodes from the top k models. The tier list was derived from having the most successful episodes. This directly reflects the models with the best clemscores.

It is to be expected that the quality of the played episodes from better models is higher than the models that only succeeded in a small number of episodes. The idea behind this experiment was to determine whether the difference in quality is reflected by the fine-tuned model.

It appears that training only on the successfully played episodes of the top 10 models has a positive impact on the quality score compared to using all available data.

\textbf{D3} In the previous experiments, a training sample consisted of a complete episode. This means, the whole conversation over multiple turns was considered as one sample. This implies that intermediate turns were not available as individual training samples in the data. This experiment was designed to determine the impact of using individual conversation pieces as training samples rather than the whole conversation at once. Therefore, every episode was split into individual continuously growing training samples that started with the first question answer pair, and then extended with each question answer pair until the end of the conversation was reached. It is important to note that the data was shuffled before splitting to ensure that conversation parts of one episode remained in the correct order and are trained on together.

We have observed that most of the experiments from D3 outperformed the respective ones from D1 and D2.

\textbf{D4} was conducted to test different balancing strategies. In the previous experiments, the dataset was not balanced between games.
Data can be balanced before or after splitting the conversation parts (as described in D3). The downsampling can be done by random selection or by considering the models' leaderboard positions. Furthermore, there can be oversampling for games with only few available episodes.
While balancing overall showed a positive impact, the best performance were achieved when the data was balanced before splitting. The sampling was based on the leaderboard without oversampling. 
This was also demonstrated by D2, where using the data from the best-performing models showed a positive impact on the fine-tuned models' performance.

\textbf{D5 \& D6} These were two complementary experiments where for D5 the model is only trained on one game, while D6 consists of the opposite experiment and can be described as leave-one-game-out. 
While this experiment did not yield meaningful insights into the dataset configuration, it led to some improvements with respect to overfitting of the probe questions in PrivateShared.

\textbf{D7 \& D8} were defined to verify or reject the possible improvements derived from game-specific data-transformations. This includes the transformations on PrivateShared and ReferenceGame.

\textbf{D9} While D1-D8 were completely focused on the data, D9 comprises a hyperparameter tuning of the QLoRA parameters.

\paragraph{Final Dataset Overview:}
\begin{table}[h]
\centering
\footnotesize
\begin{tabular}{lcc}
\hline
\textbf{Game} & \textbf{Samples Train} & \textbf{Samples Test} \\
\hline
\textbf{Before Splitting} && \\
Taboo & 434 & 18\\
Referencegame & 324 & 36\\
Wordle & 230 & 5\\
Wordle With Critic & 302 & 12\\
Wordle With Clue & 295 & 5\\
ImageGame & 278 & 12\\
PrivateShared & 214 & 5\\ 
\\
\textbf{After Splitting} && \\ 
Taboo & 560 & 22\\
ReferenceGame & 324 & 36\\
Wordle & 1038 & 19\\
Wordle With Critic & 1,192 & 105\\
Wordle With Clue & 717 & 12\\
ImageGame & 1,579 & 52\\
PrivateShared & 1,669 & 45\\ \\
Total & 7,079 & 291\\
\hline
\end{tabular}
\caption{Final Dataset composition Before and After Splitting.}
\label{tab:game_counts}
\end{table}

As shown in Table \ref{tab:game_counts}, depending on the game, the number of samples after splitting varies heavily. Eventually, the total number of samples available for the training is about 7000, while the number of samples for evaluation during the training is about 300.


\subsection{Synthetic datasets for warm start and rehearsal training regimes}
\label{appendix:synthetic_dataset}

In order to reinforce and improve the instruction-following capabilities of models during fine-tuning, we designed a synthetic dataset to use for training regimes such as warm-up training and rehearsal training.
In contrast to the data derived from clembench runs, this data consists of single-turn user-assistant interactions, and it was programmatically created with the help of human-made templates. 
These templates consist of short representations of instruction-answer-interactions which we we call `minigames'. An example of a minigame is the following:

\begin{lstlisting}[basicstyle=\small\ttfamily, breaklines=true, breakatwhitespace=true, breakindent=0pt]
[{"role": "user", "content": "Sum these numbers: 14, 26, and give the answer after  the tag SUM:"}, {"role": "assistant", "content": "SUM: 40"}]
\end{lstlisting}

We included several minigames based on letters/words, numbers, requiring the transformation of inputs into JSON format, or requiring to make choices between different options.
In minigames, the focus was on form rather than content given that we aimed through them to enhance models' capabilities of following game instructions regarding the formatting of answers or the shape of input strings. 
The final dataset consisted of around 20000 samples, obtained by filling the slots of 26 templates.
The models' answers to the minigames are required to be short and straight to the point.
Only in one case, the minigame is shaped as a multi-turn task. Here, at each turn the model has to select an item from a list only if it has not been chosen already during previous turns. The goal here was that of to enhancing attention to the overall context. The full minigame dataset is released with this work\footnote{\url{https://huggingface.co/datasets/clembench-playpen/warm-up_synthetic-data}.}.

\subsection{DPO}
DPO requires paired preference data, i.e., samples sharing the same context before positive vs negative continuations. For DPO Dialogue and DPO Turn with clembench runs, positive continuations are obtained from successful games' interactions, and negative continuations from unsuccessful and aborted games' interactions. Since the SFT models obtained top performances in \% Played for all games except Wordle and its variants, we only integrated aborted interactions for this game (and variants). Unsuccessful and aborted interactions have been collected and transformed from the same sources and with the same procedure as SFT data. For multi-player games like Taboo, data for both player 1 (i.e., giving clues) and player 2 (i.e., making guesses) have been integrated into training. The DPO Dialogue dataset consists of around 10K samples as in Fig. \ref{fig:dpo_dialogue}; the DPO Turn dataset consists of around 58K samples as in Fig. \ref{fig:dpo_turn}. 

\begin{figure*}[p]
\centering

\footnotesize
\begin{tcolorbox}[
    colback=white,
    colframe=black,
    arc=5mm,
    boxrule=1pt,
    width=\textwidth,
]
\centerline{\textbf{Prompt}}
\tcbline
You are playing a collaborative word guessing game in which you have to describe a target word for another player to guess. Rules: \\
(a) You have to reply in the form: CLUE: $<$some text$>$. Guesses from the other player will start with GUESS. \\
(b) You cannot use the target word itself, parts or morphological variants of it in your description. \\
(c) In addition, the same rules apply for related words which are provided below. \\
End conditions: \\
(i) If you use the target word or a related word in your description, then you lose. \\
(ii) If the other player can guess the target word in 3 tries, you both win. \\
Let us start. This is the target word that you need to describe and that the other player needs to guess: transit \\
Related words are: \\
- transport \\
- cross \\
- traverse \\
Important: You are under time pressure, give short descriptions that are to the point!
\end{tcolorbox}
\begin{minipage}[t]{0.49\textwidth}
\begin{tcolorbox}[
    colback=green!10,
    colframe=black,
    arc=5mm,
    boxrule=1pt,
]
\centerline{\textbf{Chosen:}}
\tcbline
\textbf{Turn 1:} \\
CLUE: Movement of people or things from one place to another. \\
GUESS: Migration \\
\textbf{Turn 2:} \\
CLUE: Typically involves a vehicle or a system of vehicles. \\
GUESS: Transportation \\
\textbf{Turn 3:} \\
CLUE: Often has fixed routes and schedules.
\end{tcolorbox}
\end{minipage}
\hfill
\begin{minipage}[t]{0.49\textwidth}
\begin{tcolorbox}[
    colback=red!10,
    colframe=black,
    arc=5mm,
    boxrule=1pt,
]
\centerline{\textbf{Rejected:}}
\tcbline
\textbf{Turn 1:} \\
CLUE: Movement of people or goods from one place to another, often using public services. \\
GUESS: Transportation \\
\textbf{Turn 2:} \\
CLUE: Not the general concept, but a specific instance or act of moving through a place, often on a scheduled route. \\
GUESS: Trip \\
\textbf{Turn 3:} \\
CLUE: Typically involves a shorter journey, often within a city or urban area, and may involve multiple stops along the way.
\end{tcolorbox}
\end{minipage}
\caption{\textbf{DPO dialogue} dataset: the initial state (prompt and game instance) is shared, the chosen and rejected continuations are the remaining turns from the successful and unsuccessful episodes.}
\label{fig:dpo_dialogue}

\scriptsize
\begin{tcolorbox}[
    colback=white,
    colframe=black,
    arc=4mm,
    boxrule=1pt,
    width=\textwidth,
]
\centerline{\textbf{Prompt}}
\tcbline
You are playing a collaborative word guessing game in which you have to describe a target word for another player to guess. Rules: \\
(a) You have to reply in the form: CLUE: $<$some text$>$. Guesses from the other player will start with GUESS. \\
(b) You cannot use the target word itself, parts or morphological variants of it in your description. \\
(c) In addition, the same rules apply for related words which are provided below. \\
End conditions: \\
(i) If you use the target word or a related word in your description, then you lose. \\
(ii) If the other player can guess the target word in 3 tries, you both win. \\
Let us start. This is the target word that you need to describe and that the other player needs to guess: independently \\
Related words are: \\
- separately \\
- alone \\
- individually \\
Important: You are under time pressure, give short descriptions that are to the point! \\
\textbf{Turn 1:} \\
CLUE: CLUE: Movement from one place to another. \\
GUESS: Travel 
\end{tcolorbox}

\begin{minipage}[t]{0.49\textwidth}
\begin{tcolorbox}[
    colback=green!10,
    colframe=black,
    arc=4mm,
    boxrule=1pt,
]
\centerline{\textbf{Chosen:}}
\tcbline
\textbf{Turn 2:} \\
CLUE: Often involves public systems like buses or trains. 
\end{tcolorbox}
\end{minipage}
\hfill
\begin{minipage}[t]{0.49\textwidth}
\begin{tcolorbox}[
    colback=red!10,
    colframe=black,
    arc=4mm,
    boxrule=1pt,
]
\centerline{\textbf{Rejected:}}
\tcbline
\textbf{Turn 2:} \\
CLUE: Not exactly, think of it as moving through a system or route. 
\end{tcolorbox}
\end{minipage}

\caption{\textbf{DPO turn} dataset: the initial state and conversational history are shared, the chosen and rejected continuations are single turns from the successful and unsuccessful episodes.}
\label{fig:dpo_turn}

\end{figure*}

For DPO Dialogue, we experiment with two variables: the number of negative samples per positive sample and the model source for negative samples. Each positive interaction is paired with \emph{n} unsuccessful – and \emph{n} aborted interactions for Wordle and variants – where \emph{n} is manipulated to find the optimal number of negative trajectories to learn from. The source for negative samples falls into three conditions: negative samples from all the models, from only the 10 best-performing models, and only models from the same family as the model to train. The best performances, as tested on clembench version \emph{1.6}, have been obtained by coupling 1 negative per positive from the 10 best-performing models. Since the first guess in wordle and wordle\_withclue is random, for DPO Turn we restrict to 2k samples from the first turn.

\begin{table}[h]
\centering
\begin{tabular}{lcc}
\hline
\textbf{Game} & \textbf{DPO Dialogue} & \textbf{DPO Turn} \\
\hline
taboo & 4.689 & 20.074\\
referencegame & 1.712 & 2.056\\
wordle\_withcritic & 1.469 & 18.234\\
wordle\_withclue & 677 & 3.380\\
imagegame & 1.291 & 12.094\\
wordle & 285 & 3.021\\
\hline
\end{tabular}
\caption{DPO datasets' samples per game.}
\label{tab:dpo_n_samples}
\end{table}

\subsection{GRPO}
\label{app_grpo_data_gen}

\paragraph{Reward Functions.}
We employed various reward functions, depending on the training game in question (see Table \ref{table_reward_fns}): these reward functions were intended to reflect the quality score computation for each game in the Clembench benchmark. Wordle (including the withclue and withcritic variants) and Referencegame used a simple 0/1 loss function: $r_g=1$ if the agent reached the correct answer within the turn limit (six and one turns, respectively), and $r_g=0$ otherwise. 

\begin{table}[h]
\centering
\begin{tabular}{|l|l|}
\hline
\textbf{Game} & \textbf{Reward Function} \\
\hline
Taboo & $(1/\textit{len}(\textit{turns}))\cdot\mathbb{I}_\textit{corr}$ \\
\hline
Wordle & \\
(+withclue, & $\mathbb{I}_\textit{corr}$ \\
+withcritic) & \\
\hline
Referencegame & $\mathbb{I}_\textit{corr}$ \\
\hline
Imagegame & $F_1(G_\textit{pred},G_\textit{target})$ \\
\hline
Privateshared & $\textit{Acc}(S_\textit{pred},S_\textit{target})$ \\
\hline
\end{tabular}
\caption{GRPO reward functions by training game. $\mathbb{I}_\textit{corr}=1$ if the agent reached the correct answer within the maximum number of turns, and $\mathbb{I}_\textit{corr}=0$ otherwise.}
\label{table_reward_fns}
\end{table}

For Imagegame, $r_g$ is given by the $F_1$ score between the agent's predicted grid and the target grid, and $r_g$ is given by the agent's slot-filling accuracy for privateshared. The reward function for Taboo incorporates a length penalty: $r_g=1/n$ if the agent guesses the correct word within $n\leq3$ turns, and $r_g=0$ otherwise.

For all seven games, the $r_g=0$ if the game was aborted due to agent error, such as incorrect formatting or violation of the game's rules (e.g.\hspace{1mm}predicting a non-five-letter word in Wordle).

\paragraph{Teacher Model.}
In the offline learning experiments conducted in this paper (Section \ref{subsubsec:imitation_learning} and Section \ref{subsubsec:direct_alignment}), the models were trained to play both roles in all two-player games: this was not practical for the online RL experiment. If the model is trained in both roles, approximation of the optimal policy is intractable: updates to the current policy are dependent on the reward, which is dependent on the environment, which in turn is dependent on the current policy (via the generations of the current policy playing in the other role).

On the other hand, if we employ a \textit{different}, frozen model as the second player (the \textit{teacher} model), approximation of the optimal policy becomes tractable. However, evaluating the agent model against itself (i.e.\hspace{1mm}playing both roles) introduces a mismatch between the train and test splits: the agent will have approximated the optimal policy for the environment in which the teacher is the second player.

Despite this train-test mismatch, we trained the GRPO agent with GPT-4o-mini\footnote{\url{https://platform.openai.com/docs/models/gpt-4o-mini}} as the teacher model: in line with our goal of \textit{learning} from interaction, we employed a more advanced model than our agent (Llama-3.1-8B), to enable the agent to learn from its teacher/caregiver. A list of the two-player games\textemdash and the roles played by the teacher and agent in each\textemdash is given in Table \ref{table_agent_teacher_roles}.

\paragraph{Challenges and Adaptations of the Playpen Environment.}
We adapted the Playpen environment to online RL applications by re-configuring Playpen to allow individual game instances to be played separately: this allows for the tuning of batch size as a hyperparameter, and the random permutation of game instances across batches. We additionally implemented non-agent token masking, so that the agent's loss is only computed with respect to its own generated tokens.

\subparagraph{Teacher-Aborted Episodes}
In the Playpen environment, an episode can be aborted if there is a rule violation from either the agent or teacher model: for example, if the teacher model includes the target word in its clue during a Taboo game instance. 

In the case of teacher error, the agent model should not be negatively rewarded due to the aborted episode. To account for this, we set a \textit{retry limit} $\rho$, such that a teacher-aborted episode will be replayed up to $\rho$ times in the case of teacher error\footnote{In practice, we set $\rho=1$ for all experiments due to computational resource limitations.}. 

If a single instance $g\in G_i$ has been aborted $\rho$ times due to teacher error, we replace $g$ with another randomly-selected $g'\neq g\in G_i$ from the same group for loss computation and backpropagation. If \textit{every} instance in the group $G_i$ is aborted $\rho$ times due to teacher error, we replace $G_i$ with another group $G_{k\neq i}$ in the same batch. 

\subparagraph{Privateshared}
The privateshared game was particularly problematic for online RL, as the quality score for this game is primarily computed from probes that are conducted adjacent to the actual game, and the transcripts from these probes are removed from the agent's observations after they are completed. 

Including the probes in the instance trajectory during training results in a mismatch between the train and test splits, as the agent only sees the \textit{current} probe at test time. Conversely, removing the probes from the trajectory leads to unpredictable rewards from the environment: if the agent's reward is negatively affected by its performance in a probe, the reason for the negative reward will not be reflected in the trajectory.

For these reasons, we did not consider the agent's probing-task performance in the computation of the privateshared reward function. This has a severe negative effect on test-set performance for this game: online RL substantially degrades the model's quality score for Privateshared (see Appendix \ref{app_results_grpo}), even when beginning online RL from the best SFT Llama model. 

\section{Training details} \label{app:training_details}

\subsection{Supervised Fine-tuning}
\label{appendix:training_details_sft}

The SFT models are fine-tuned using QLoRA \citep{detmers:qlora} adapters ($r=64$, $\alpha=32$, $\textit{dropout}=0.05$) on all linear layers. The models were trained with the following arguments ($optim=adamw\_8bit$, $lr=2e-4$, $lr \,scheduler=linear$, $decay=0.01$, $batch \,size=4$, $steps=600-700$ and fixed $seed=7331$). The models were quantized in 4-bit using the \textit{unsloth} \citep{unsloth} library and the following bits-and-bytes configuration ($use\_4bit=True$, $bnb\_4bit\_compute\_dtype=float16$, $bnb\_4bit\_quant\_type=nf4$, $use\_nested\_quant=False$). As a stopping criterion, the first checkpoint before the minimal evaluation loss that has a distance of less than or equal to 0.015 from the minimal evaluation loss was chosen. Hence, a full epoch must be trained to determine the optimal checkpoint. 
The most relevant libraries and their versions are ($torch=2.4.0$, $unsloth=2024.8$, $transformers=4.47.1$, $bitsandbytes=0.43.3$, $trl=0.9.6$, $accelerate=0.34.2$). 

\paragraph{Training Setup}
All previously described experiments were conducted on a quantized version of Llama-3.1-8B (Instruct version).
All models were fine-tuned using Unsloth \cite{unsloth} with 4-bit quantization and QLoRA \cite{detmers:qlora} for a more efficient and resource-optimized fine-tuning.


\paragraph{Hardware} The training was conducted on one NVIDIA A100 GPU with 80 GB of VRAM and one NVIDIA H100 GPU with 95 GB of VRAM. It should be noted that technically a multi-GPU setup was possible, but every experiment was only conducted on a single-GPU. 

\paragraph{Training Procedure}
In the first step, the models were trained on all available training data. Based on the training statistics (train and evaluation loss), a second model was trained using the number of steps with the lowest evaluation loss.

To address the issue of over-fitting, a third model was trained using significantly fewer steps. The number of steps was chosen based on the evaluation loss, with a threshold set to 0.015. The third model was trained until the evaluation loss reached a value within or equal to this threshold relative to the best evaluation loss.

\textbf{As an example:}
The first model is trained for 1700 steps (all available data), but the minimal evaluation loss is reached at around 1100 steps. With a minimal eval-loss of 0.2315, the second model is trained for 1100 steps while the third model is trained until the eval-loss reaches the threshold of 0.2315 + 0.015 = 0.2430.
The final model required 700 training steps. This approach helps to prevent over-fitting, as continuing training beyond the threshold (where evaluation loss increases by 0.015) provides diminishing returns while potentially reducing generalization capabilities. This is a strategy we call \texttt{SFT (Cold Start, CS)} in the main paper. 

We also experimented with other variants that we report below:
\begin{itemize}
\item \texttt{SFT (Warm Start, WS)}: Before training on the interaction data, the model was trained on 100 steps (400 samples) of synthetic instruction following tasks, using the findings from \cite{gupta2023warmstart} and focusing on instruction following abilities. 
\item \texttt{SFT (Rehearsal, R)}: during training, we interleave the gameplay training dataset with basic instruction following data following a similar approach to \citet{lambert_tulu_2025}. 
\end{itemize}





\subsection{DPO}
For both DPO Dialogue and DPO Turn, an SFT QLoRA adapter has been mounted on top of the base model Llama3.1-8B. To merge the base model and the SFT adapter, three merging strategies have been tested before DPO: merging the full-precision Llama3.1-8B model with the adapter, merging the unsloth 4-bit quantized Llama3.1-8B version with the adapter in 16bit, and merging the unsloth 4-bit quantized Llama3.1-8B with the adapter in 4bit. As reported in Tab. \ref{tab:appendix_merging_strategies}, the first strategy outperforms the others, showing comparable results to the unmerged adapter.

\begin{table}[h!]
\centering
\footnotesize
\begin{tabular}{lrrr}
\toprule
\textbf{Model} & \textbf{ClemScore} & \textbf{pp} & \textbf{qs} \\
\midrule
unmerged & 46.82 & 75.24 & 62.23 \\
full-precision & 47.79 & 74.88 & 63.82 \\
16bit & 33.52 & 70.19 & 47.76 \\
4bit & 30.14 & 60.00 & 50.23 \\
\bottomrule
\end{tabular}
\caption{Comparison of merging strategies in terms of Clemscore, average \% played (pp) and quality score (qs).}\label{tab:appendix_merging_strategies}
\end{table}

DPO training is performed on top of the 4-bit quantized SFT model, with the same bits-and-bytes configuration as the SFT models ($use\_4bit=True$, $bnb\_4bit\_compute\_dtype=float16$, $bnb\_4bit\_quant\_type=nf4$, $use\_nested\_quant=False$). QLoRA adapters are employed on the same modules as for SFT (with $r=64$, $\alpha=64$, and $\textit{dropout}=0$). The models have been trained with the $adamw\_8bit$ optimizer, a learning rate of $5e-6$, with $linear$ lr scheduler and the $beta=0.1$ ($decay=0$, $batch \,size=2$, $gradient \,accumulation\,steps = 3$ and fixed $seed=42$). During training, we evaluate the model every 20\% on held-out training samples. At the end of training, only the best-performing checkpoints on the dev sets were saved. The relevant libraries' versions are: $torch=2.5.1$, $unsloth=2024.12.4$, $transformers=4.46.3$, $bitsandbytes=0.45.0$, $trl=0.12.2$, $accelerate=1.2.0$.

In terms of hardware, DPO development has been performed on 2xA5000s. Large differences have been observed when comparing results obtained on the A5000 and A100. The final training for clembench \emph{v2.0} has been performed on an RTX3090 with 24GB RAM. 

\subsection{GRPO}
We conducted two online RL experiments: one pure RL experiment, in which we initialized the agent from the baseline Llama-3.1-8B model (Instruct version) (GRPO); and a second experiment in which the RL agent was initialized from the best-performing SFT model (SFT(CS)+GRPO).

The training set for both experiments consisted of game instances from Clembench V0.9 and V1.0 for Taboo (90 instances), Wordle (60), Wordle-withclue (60), Wordle-withcritic (60), Referencegame (256), Imagegame (80), and Privateshared (80), for a total of 686 instances. The validation split consisted of 420 Clembench V1.6 game instances (total) for the training games. 

Both GRPO models were trained on four NVIDIA H100 GPUs with 80 GB of VRAM (each): for speedup, trajectory generation was parallelized across the four GPUs.

Both GRPO models were tuned using LoRA \citep{hu2022lora} adapters ($r=64$, $\alpha=128$, $\textit{dropout}=0$) on their $Q$, $K$, $V$, and $O$ attention projection matrices. We trained the models for five epochs on 686 game instances with a temperature of 0.75, a batch size of 16, a group size of 8, KL regularization $\beta=0.04$, and a learn rate of $10^{-6}$ using the Adam optimizer (for GRPO $seed = 250329152534053703$, for SFT(CS)+GRPO $seed = 250327114458100881$).

\subsection{Evaluation Details}
The code for performing the experiments on the non-interactive benchmarks is available here: \url{https://github.com/momentino/playpen_eval/tree/playpen}.

The evaluation of the models on non-interactive datasets was conducted on Ampere-architecture GPUs (A100, A40). The experiments have been conducted by extending the \textit{lm-eval} framework with the tasks which were not present in its original version (i.e. CLadder, LM-Pragmatics, \textsc{Natural Plan}, GLUE Diagnostics, SimpleToM). Out of these, CLadder, \textsc{Natural Plan} and SimpleToM have been taken without any modification from those implemented by \cite{momente2025triangulating}. \textsc{Natural Plan} has also been taken from there, but we removed the upper and lower bounds on the number of tokens that the model is allowed to generate. GLUE Diagnostics was implemented from scratch.
For ensuring a more efficient evaluation, we relied on the vLLM library \cite{vllm}  (version 0.8.3).

To ensure comparability of the results, all evaluations on clembench v2.0 were carried out exclusively on an H100 GPU. It appears that when using different GPUs, the results can differ by up to 5 percentage points in some models. The H100 was chosen due to its higher inference speeds to save time on evaluation. 





\section{Results}
\label{app:additional_results}

\subsection{Supervised Fine-tuning}


\begin{table*}[h!]
\centering
\resizebox{0.8\textwidth}{!}{%
\begin{tabular}{lcccc}
\toprule
\textbf{Game/Model}     & \textbf{L3-8B (Baseline)} & \textbf{PP L3-8B CS}                                           & \textbf{PP L3-8B WS} & \textbf{PP L3-8B R} \\ \midrule
\textit{\textbf{In Domain}}      & pp/qs          & pp/qs                                                       & pp/qs   & pp/qs  \\
\textbf{ImageGame}      & 67.8/54.62     & {\color[HTML]{38761D} 32.20}/{\color[HTML]{38761D} 39.87}   &   {\color[HTML]{38761D} 32.20}/{\color[HTML]{38761D} 37.65}  &   {\color[HTML]{38761D} 32.20}/{\color[HTML]{38761D} 39.19}\\
\textbf{PrivateShared}     & 100/23.48      & 0.00/{\color[HTML]{38761D} 73.65}                           &   0.00/{\color[HTML]{38761D} 73.05}  &   {\color[HTML]{990000} -4.00}/{\color[HTML]{38761D} 69.16}\\
\textbf{ReferenceGame}  & 100/38.89      & 0.00/{\color[HTML]{38761D} 7.78}                            &  0.00/{\color[HTML]{990000} -3.33}  &   0.00/{\color[HTML]{990000} -4.45}\\
\textbf{Taboo}          & 98.33/31.92    & {\color[HTML]{38761D} 1.67}/{\color[HTML]{38761D} 5.58}     &   {\color[HTML]{38761D} 1.67}/{\color[HTML]{38761D} 3.91}  &   {\color[HTML]{38761D} 1.67}/{\color[HTML]{38761D} 9.19}\\
\textbf{Wordle}         & 36.67/0        & {\color[HTML]{38761D} 20.00}/{\color[HTML]{38761D} 1.18}    &   {\color[HTML]{38761D} 30.00}/{\color[HTML]{38761D} 5.00}  &   {\color[HTML]{38761D} 16.66}/{\color[HTML]{38761D} 8.12}\\
\textbf{Wordle With Clue}    & 0/-            & {\color[HTML]{38761D} 23.33}/{\color[HTML]{38761D} 71.43}   &  {\color[HTML]{38761D} 10.00}/{\color[HTML]{38761D} 16.67}  &   {\color[HTML]{38761D} 6.67}/0.00\\
\textbf{Wordle With Critic}    & 6.67/50        & {\color[HTML]{38761D} 6.66}/0.00                            &  {\color[HTML]{990000} -3.34}/{\color[HTML]{38761D} 50.00}  &   {\color[HTML]{990000} -3.34}/{\color[HTML]{990000} -16.67}\\ \midrule
\textit{\textbf{Out-of-Domain}}  & pp/qs          & pp/qs                                                       & pp/qs   & pp/qs  \\
\textbf{Adventure Game}       & 35.94/33.85    & {\color[HTML]{990000} -17.97}/{\color[HTML]{990000} -18.23} &   {\color[HTML]{990000} -15.63}/{\color[HTML]{990000} -26.93}  &   {\color[HTML]{990000} -15.36}/{\color[HTML]{990000} -17.70}\\
\textbf{Codenames}      & 43.08/16.07    & {\color[HTML]{990000} -17.70}/{\color[HTML]{38761D} 5.14}   &   {\color[HTML]{990000} -26.93}/{\color[HTML]{38761D} 17.26}  &   {\color[HTML]{990000}-17.7}/{\color[HTML]{38761D}5.14}\\
\textbf{Map Navigation}            & 36/55.46       & {\color[HTML]{38761D} 32.00}/{\color[HTML]{990000} -8.75}   &   {\color[HTML]{38761D} 24.00}/{\color[HTML]{990000} -2.31}  &   {\color[HTML]{38761D} 32.00}/{\color[HTML]{990000} -0.11}\\
\textbf{Map Navigation (Graph)}     & 20/44.33       & {\color[HTML]{990000} -3.33}/{\color[HTML]{990000} -7.54}   &   {\color[HTML]{990000} -16.67}/{\color[HTML]{990000} -15.76}  &   {\color[HTML]{990000} -13.33}/{\color[HTML]{990000} -13.38}\\
\textbf{Map Navigation (Room)}      & 56.67/94.12    & {\color[HTML]{990000} -16.67}/{\color[HTML]{990000} -2.45}  &   {\color[HTML]{990000} -6.67}/{\color[HTML]{990000} -7.45}  &   {\color[HTML]{38761D} 16.66}/{\color[HTML]{990000} -16.85}\\
\textbf{MatchIt (ASCII)} & 100/60         & {\color[HTML]{990000} -2.50}/{\color[HTML]{38761D} 9.23}    &   {\color[HTML]{990000} -10.00}/{\color[HTML]{990000} -26.67}  &   0.00/{\color[HTML]{38761D} 7.50}\\
\textbf{GuessWhat}      & 90/11.73       & {\color[HTML]{990000} -1.67}/{\color[HTML]{38761D} 19.09}   &  {\color[HTML]{990000} -18.33}/{\color[HTML]{38761D} 20.05}  &   {\color[HTML]{990000} -13.33}/{\color[HTML]{990000} -3.67}\\ \bottomrule
\end{tabular}%
}
\caption{Gains and losses w.r.t baselines of average \% played and quality score of individual games; L3: Llama-3.1-8B-Instruct, PP: Playpen, CS: Cold Start, WS: Warm Start, R: Rehearsal.}
\label{tab:GameSpecificResultsChangeLlama8B}
\end{table*}

Table \ref{tab:GameSpecificResultsChangeLlama8B} depicts results achieved by the Llama-3.1-8B-Instruct baseline and the three variants Cold Start (CS), Warm Start (WS), and Rehearsal (R) on games from \texttt{clembench} 2.0. The upper half of the table shows results for in-domain games while the lower half those for out-of-domain games.

It becomes visible that in-domain the three fine-tuned models appear to have quite substantial performance gains compared to the baseline, while for out-of-domain games the opposite is true. 

Comparing the three fine-tuning versions it appears that, overall, the cold start one outperforms the other two. While WS and R show slight decreases in PrivateShared and ReferenceGame (in-domain games), the CS version displays improvements for all in-domain games.

Regarding out-of-domain games, the performance in- and decreases shifted between models but some patterns (e.g. Codenames and Map Navigation improvements) still remain. A larger discrepancy can be seen for Map Navigation (Graph). Here, WS and R seem to negatively impact the model performance. Map Navigation (Graph) is the only game that requires the model to produce a valid JSON-object. In this game in particular it is crucial to follow a strict output format since a malformed JSON leads to an aborted game.

Interestingly, Llama-3.1-8B (CS) shows out-of-domain a more substantial decrease in the \% played score compared to quality score. For the other two models it appears that the performance loss is more balanced across the two scores. For Llama-3.1-8B (CS) this indicates that the fine-tuning negatively impacts the models ability to properly play the game. The \% played is an indicator of what \% of episodes were actually played and how many were aborted. This is tightly bound to game-specific output formats especially for the in-domain games. As for the out-of-domain games, the played score sometimes will be also negatively impacted if the model reaches a turn-limit. Even though the model knows how to structure the output, the episode will be counted as aborted. This makes it difficult to pinpoint the exact reason for the decrease in \% played to one particular cause. The reason may be caused by an over-fitting to the prompt structure of the in-domain games-specific instructions. Alternatively, it could also be that other abilities such as the contextual awareness of the model are worsened by the fine-tuning process, and this may lead to reach the turn limit more easily.

\subsection{DPO}
Detailed results are provided in Table  \ref{tab:dpo_res}. Compared to the base L3-8B SFT(CS), both DPO Dialogue and DPO turn appear to result in a degradation of performance on in-domain games, with the most pronounced declines observed for Wordle. On the other hand, in out-of-domain games, improvements in many games are observed for DPO Dialogue, with peaks in Map Navigation (Graph) and Map Navigation (Room). DPO Turn, instead, seems to perform worse than the baseline model for most of the games.

\begin{table*}[h!]
\centering
\resizebox{0.8\textwidth}{!}{%
\begin{tabular}{lcccc}
\toprule
\textbf{Game/Model} & \textbf{L3-8B SFT(CS)} & \textbf{SFT(CS)+DPO Dialogue} & \textbf{SFT(CS)+DPO Turn} \\
\midrule
\textit{\textbf{In Domain}} & pp/qs & pp/qs & pp/qs \\
\textbf{ImageGame} & 100/94.49 & {\color[HTML]{990000} -15.25}/{\color[HTML]{990000} -15.07} & 0.0/{\color[HTML]{990000} -1.93} \\
\textbf{PrivateShared} & 100/97.13 & 0.0/{\color[HTML]{990000} -4.40} & 0.0/{\color[HTML]{990000} -1.28} \\
\textbf{ReferenceGame} & 100/46.67 & 0.0/{\color[HTML]{990000} -4.45} & 0.0/{\color[HTML]{990000} -5.56} \\
\textbf{Taboo} & 100/37.5 & 0.0/{\color[HTML]{38761D} 6.94} & 0.0/{\color[HTML]{38761D} 7.22} \\
\textbf{Wordle} & 56.67/1.18 & {\color[HTML]{990000} -30.0}/{\color[HTML]{38761D} 2.57} & {\color[HTML]{990000} -13.34}/{\color[HTML]{38761D} 1.13} \\
\textbf{Wordle With Clue} & 23.33/71.43 & 0.0/{\color[HTML]{990000} -21.43} & {\color[HTML]{38761D} 10.0}/{\color[HTML]{990000} -48.93} \\
\textbf{Wordle With Critic} & 26.66/27.78 & {\color[HTML]{990000} -3.33}/{\color[HTML]{38761D} 2.78} & {\color[HTML]{38761D} 13.34}/{\color[HTML]{990000} -25.0} \\
\midrule
\textit{\textbf{Out-of-Domain}} & pp/qs & pp/qs & pp/qs \\
\textbf{Adventure Game} & 17.97/15.62 & {\color[HTML]{38761D} 2.34}/{\color[HTML]{990000} -3.9} & {\color[HTML]{990000} -3.91}/0.0 \\
\textbf{Codenames} & 25.38/21.21 & {\color[HTML]{990000} -15.38}/{\color[HTML]{990000} -5.83} & {\color[HTML]{990000} -8.46}/{\color[HTML]{990000} -7.57} \\
\textbf{Map Navigation} & 68/46.71 & {\color[HTML]{38761D} 12.0}/{\color[HTML]{38761D} 9.0} & {\color[HTML]{990000} -8.0}/{\color[HTML]{38761D} 5.28} \\
\textbf{Map Navigation (Graph)} & 16.67/36.79 & {\color[HTML]{38761D} 43.33}/{\color[HTML]{38761D} 9.18} & {\color[HTML]{990000} -3.34}/{\color[HTML]{990000} -0.91} \\
\textbf{Map Navigation (Room)} & 40/91.67 & {\color[HTML]{38761D} 23.33}/{\color[HTML]{990000} -7.46} & {\color[HTML]{38761D} 8.33}/{\color[HTML]{990000} -13.34} \\
\textbf{MatchIt (ASCII)} & 97.5/69.23 & {\color[HTML]{990000} -70.0}/{\color[HTML]{990000} -51.05} & 0.0/{\color[HTML]{990000} -10.26} \\
\textbf{GuessWhat} & 88.33/30.82 & {\color[HTML]{38761D} 6.67}/{\color[HTML]{990000} -3.92} & {\color[HTML]{38761D} 10.0}/{\color[HTML]{990000} -4.27} \\
\bottomrule
\end{tabular}%
}
\caption{Comparison of \% played (pp) and quality score (qs) on individual games for the L3-8B SFT(CS) and the further trained DPO Dialogue and DPO Turn.}
\label{tab:dpo_res}
\end{table*}

\subsection{GRPO}
\label{app_results_grpo}

\begin{table}[h!]
\centering
\scriptsize
\begin{tabular}{@{}p{2.4cm}|p{2.4cm}|p{2.0cm}@{}}
\textbf{Game} & \textbf{Agent Role} & \textbf{Teacher Role} \\
\hline
ImageGame & Instruction Follower & Instruction Giver \\
\hline
ReferenceGame & Instruction Follower & Instruction Giver \\
\hline
Taboo & Guesser & Describer \\
\hline
Wordle With Critic & Guesser & Critic \\
\hline
\end{tabular}
\caption{Two-player games from the train split, and the roles played by the agent and teacher models in each for the online RL experiment.}
\label{table_agent_teacher_roles}
\vspace{-1em}
\end{table}

The performance increases and decreases for the GRPO models relative to their respective baselines are given in Table \ref{tab:table_grpo_res}. Pure reinforcement learning leads to near-across-the-board improvements over the baseline Llama 3.1 8B model on all in- and out-of-domain games, although we observe slight decreases in percentage played on Wordle, Wordle With Critic, Map Navigation, Map Navigation (Room), and GuessWhat, along with slight decreases in quality score for the latter three out-of-domain games.

On the other hand, GRPO struggles to improve the SFT Llama model (SFT(CS)+GRPO), and only results in slight increases in quality score for Taboo, Wordle, and Wordle With Critic. We also observe substantial decreases in in-domain performance, in particular on ImageGame and PrivateShared: the decrease in PrivateShared is to be expected, as the reward function for this game is only loosely connected to the \texttt{clemscore} (as discussed in Appendix \ref{app_grpo_data_gen}). However, GRPO greatly improves the out-of-domain clemscores of the SFT model\textemdash with the notable exceptions of MatchIt (ASCII) and GuessWhat. 

\begin{table*}[h!]
\centering
\resizebox{0.85\textwidth}{!}{%
\begin{tabular}{lcccc}
\toprule
\textbf{Game/Model} & \textbf{L3-8B} & \textbf{GRPO} & \textbf{L3-8B SFT(CS)} & \textbf{SFT(CS)+GRPO} \\
\midrule
\textit{\textbf{In Domain}} & pp/qs & pp/qs & pp/qs & pp/qs \\
\textbf{ImageGame} & 67.8/54.62 & {\color[HTML]{38761D} 1.69}/{\color[HTML]{38761D} 2.65} & 100/94.49 & {\color[HTML]{990000} -24.14}/{\color[HTML]{990000} -43.38} \\
\textbf{PrivateShared} & 100/23.48 & 0.0/{\color[HTML]{38761D} 0.69} & 100/97.13 & 0.0/{\color[HTML]{990000} -76.17} \\
\textbf{ReferenceGame} & 100/38.89 & 0.0/{\color[HTML]{38761D} 4.44} & 100/46.67 & 0.0/{\color[HTML]{990000} -10.0} \\
\textbf{Taboo} & 98.33/31.92 & {\color[HTML]{38761D} 1.67}/{\color[HTML]{38761D} 2.8} & 100/37.5 & 0.0/{\color[HTML]{38761D} 4.39} \\
\textbf{Wordle} & 36.67/0 & {\color[HTML]{990000} -6.37}/0.0 & 56.67/1.18 & {\color[HTML]{990000} -3.34}/{\color[HTML]{38761D} 1.94} \\
\textbf{Wordle With Clue} & 0/- & 0.0/- & 23.33/71.43 & {\color[HTML]{990000} -20.0}/{\color[HTML]{990000} -21.43} \\
\textbf{Wordle With Critic} & 6.67/50 & {\color[HTML]{990000} -3.34}/{\color[HTML]{38761D} 50.0} & 26.66/27.78 & {\color[HTML]{990000} -16.66}/{\color[HTML]{38761D} 50.0} \\
\midrule
\textit{\textbf{Out-of-Domain}} & pp/qs & pp/qs & pp/qs & pp/qs \\
\textbf{Adventure Game} & 35.94/33.85 & {\color[HTML]{38761D} 19.62}/{\color[HTML]{38761D} 11.28} & 17.97/15.62 & {\color[HTML]{38761D} 28.7}/{\color[HTML]{38761D} 20.82} \\
\textbf{Codenames} & 43.08/16.07 & {\color[HTML]{38761D} 3.84}/{\color[HTML]{38761D} 15.08} & 25.38/21.21 & {\color[HTML]{38761D} 13.85}/{\color[HTML]{990000} -3.56} \\
\textbf{Map Navigation} & 36/55.46 & {\color[HTML]{990000} -6.0}/{\color[HTML]{990000} -1.35} & 68/46.71 & {\color[HTML]{38761D} 6.0}/{\color[HTML]{38761D} 13.37} \\
\textbf{Map Navigation (Graph)} & 20/44.33 & {\color[HTML]{38761D} 57.78}/{\color[HTML]{38761D} 1.36} & 16.67/36.79 & {\color[HTML]{38761D} 64.28}/{\color[HTML]{38761D} 13.83} \\
\textbf{Map Navigation (Room)} & 56.67/94.12 & {\color[HTML]{990000} -3.34}/{\color[HTML]{990000} -0.37} & 40/91.67 & {\color[HTML]{38761D} 10.0}/{\color[HTML]{38761D} 1.66} \\
\textbf{MatchIt (ASCII)} & 100/60 & 0.0/{\color[HTML]{38761D} 2.5} & 97.5/69.23 & {\color[HTML]{38761D} 2.5}/{\color[HTML]{990000} -6.73} \\
\textbf{GuessWhat} & 90/11.73 & {\color[HTML]{990000} -10.0}/{\color[HTML]{990000} -1.31} & 88.33/30.82 & {\color[HTML]{990000} -8.33}/{\color[HTML]{990000} -20.4} \\
\bottomrule
\end{tabular}%
}
\caption{Gains and losses w.r.t baseline of average \% played (pp) and quality score (qs) of individual games.}
\label{tab:table_grpo_res}
\end{table*}

\subsection{Evaluation on General Instruction Following Benchmarks}
\label{sec:appendix-eval-instruction-following}
We report in Table~\ref{tab:reference-based-results-detailed-1} and Table~\ref{tab:reference-based-results-detailed-2} a detailed breakdown of the results obtained in the evaluation on general instruction following tasks considered in this study. 

\begin{table*}[ht]
\centering
\resizebox{\textwidth}{!}{%
\begin{tabular}{l
llll|  
lllll  
}
\toprule
& \multicolumn{4}{c}{\textbf{Executive}} 
& \multicolumn{5}{c}{\textbf{Socio-Emotional}} \\
 & \textsc{Natural Plan}
& \textit{LogiQA2} & \textit{CLadder} & \textit{WinoGrande} 
& \textit{EQBench} & \textit{LM-Pragmatics} & \textit{SocialIQA} & \textit{SimpleToM (AJ)} & \textit{SimpleToM (ToM)} \\

\midrule
\textbf{Llama-3.1-8B} &&&&&&&&\\

Base & 06.40 & 32.31 & 50.57 & 67.71 & 67.79 & 65.12 & 48.36 & 46.68 & 57.71 \\

SFT (CS) & 09.83 {\scriptsize \textcolor{green!60!black}{\rule{3.43pt}{5pt}}} & 31.11 {\scriptsize \textcolor{red}{\rule{1.2pt}{5pt}}} & 54.13 {\scriptsize \textcolor{green!60!black}{\rule{6.43pt}{3.56pt}}} & 64.64 {\scriptsize \textcolor{red}{\rule{3.07pt}{5pt}}} & 61.45 {\scriptsize \textcolor{red}{\rule{6.34pt}{5pt}}} & 62.08 {\scriptsize \textcolor{red}{\rule{3.04pt}{5pt}}} & 47.24 {\scriptsize \textcolor{red}{\rule{1.12pt}{5pt}}} & 38.45 {\scriptsize \textcolor{red}{\rule{8.23pt}{5pt}}} & 88.31 {\scriptsize \textcolor{green!60!black}{\rule{3.06pt}{5pt}}} \\

SFT (WS) & 06.17 {\scriptsize \textcolor{red}{\rule{0.23pt}{5pt}}} & 32.63 {\scriptsize \textcolor{green!60!black}{\rule{0.32pt}{5pt}}} & 51.95 {\scriptsize \textcolor{green!60!black}{\rule{1.38pt}{5pt}}} & 69.69 {\scriptsize \textcolor{green!60!black}{\rule{1.98pt}{5pt}}} & 51.72 {\scriptsize \textcolor{red}{\rule{16.07pt}{5pt}}} & 45.73 {\scriptsize \textcolor{red}{\rule{19.39pt}{5pt}}} & 47.85 {\scriptsize \textcolor{red}{\rule{0.51pt}{5pt}}} & 34.26 {\scriptsize \textcolor{red}{\rule{12.42pt}{5pt}}} & 71.83 {\scriptsize \textcolor{green!60!black}{\rule{14.12pt}{5pt}}} \\

SFT (R) & 12.80 {\scriptsize \textcolor{green!60!black}{\rule{6.4pt}{5pt}}} & 32.18  {\scriptsize \textcolor{red}{\rule{0.13pt}{5pt}}} & 53.37 {\scriptsize \textcolor{green!60!black}{\rule{2.8pt}{5pt}}} & 67.88 {\scriptsize \textcolor{green!60!black}{\rule{0.17pt}{5pt}}} & 49.84 {\scriptsize \textcolor{red}{\rule{17.95pt}{5pt}}} & 55.97 {\scriptsize \textcolor{red}{\rule{9.15pt}{5pt}}} & 49.84 {\scriptsize \textcolor{green!60!black}{\rule{1.48pt}{5pt}}} & 24.80 {\scriptsize \textcolor{red}{\rule{21.88pt}{5pt}}} & 82.39 {\scriptsize \textcolor{green!60!black}{\rule{24.68pt}{5pt}}} \\

SFT (CS) + DPO (Dial.) & 12.17 {\scriptsize \textcolor{green!60!black}{\rule{5.77pt}{5pt}}} & 27.16 {\scriptsize \textcolor{red}{\rule{5.15pt}{5pt}}} & 53.36 {\scriptsize \textcolor{green!60!black}{\rule{2.79pt}{5pt}}} & 61.09 {\scriptsize \textcolor{red}{\rule{6.62pt}{5pt}}} & 61.48 {\scriptsize \textcolor{red}{\rule{6.31pt}{5pt}}} & 50.85 {\scriptsize \textcolor{red}{\rule{14.27pt}{5pt}}} & 44.88 {\scriptsize \textcolor{red}{\rule{3.48pt}{5pt}}} & 38.19 {\scriptsize \textcolor{red}{\rule{8.49pt}{5pt}}} & 80.12 {\scriptsize \textcolor{green!60!black}{\rule{22.41pt}{5pt}}} \\

SFT (CS) + DPO (Turn) & 13.33 {\scriptsize \textcolor{green!60!black}{\rule{6.93pt}{5pt}}}  & 32.32 {\scriptsize \textcolor{green!60!black}{\rule{0.01pt}{5pt}}}  & 51.97 {\scriptsize \textcolor{green!60!black}{\rule{1.4pt}{5pt}}}  & 61.96 {\scriptsize \textcolor{red}{\rule{5.75pt}{5pt}}}  & 60.22 {\scriptsize \textcolor{red}{\rule{7.57pt}{5pt}}}  & 61.60 {\scriptsize \textcolor{red}{\rule{3.52pt}{5pt}}}  & 47.59 {\scriptsize \textcolor{red}{\rule{0.77pt}{5pt}}}  & 41.37 {\scriptsize \textcolor{red}{\rule{5.31pt}{5pt}}}  & 87.97 {\scriptsize \textcolor{green!60!black}{\rule{30.26pt}{5pt}}}  \\

GRPO & 07.31 {\scriptsize \textcolor{green!60!black}{\rule{0.91pt}{5pt}}} & 32.12 {\scriptsize \textcolor{red}{\rule{0.19pt}{5pt}}} & 50.96 {\scriptsize \textcolor{green!60!black}{\rule{0.39pt}{5pt}}}  & 67.17 {\scriptsize \textcolor{red}{\rule{0.54pt}{5pt}}} & 67.69  {\scriptsize \textcolor{red}{\rule{0.1pt}{5pt}}} & 65.49 {\scriptsize \textcolor{green!60!black}{\rule{0.37pt}{5pt}}} & 48.56 {\scriptsize \textcolor{green!60!black}{\rule{0.2pt}{5pt}}} & 46.60 {\scriptsize \textcolor{red}{\rule{0.08pt}{5pt}}} & 59.20 {\scriptsize \textcolor{red}{\rule{1.49pt}{5pt}}} \\

SFT (CS) + GRPO & 05.42 {\scriptsize \textcolor{red}{\rule{0.98pt}{5pt}}} & 32.06 {\scriptsize \textcolor{red}{\rule{0.25pt}{5pt}}} & 29.24 {\scriptsize \textcolor{red}{\rule{21.33pt}{5pt}}} & 66.69  {\scriptsize \textcolor{red}{\rule{1.02pt}{5pt}}} & 68.94 {\scriptsize \textcolor{green!60!black}{\rule{1.15pt}{5pt}}} & 65.61 {\scriptsize \textcolor{green!60!black}{\rule{0.49pt}{5pt}}} & 48.56 {\scriptsize \textcolor{green!60!black}{\rule{0.2pt}{5pt}}} & 50.04 {\scriptsize \textcolor{green!60!black}{\rule{3.36pt}{5pt}}} & 60.24 {\scriptsize \textcolor{green!60!black}{\rule{2.53pt}{5pt}}} \\ 

&&&&&&&&&\\
\textbf{Llama-3.1-70B} &&&&&&&&\\

Base & 29.03 & 51.52 & 56.34 & 72.77 & 82.03 & 80.97 &  55.02 & 44.33 & 94.5\\

SFT(CS) & 32.03 {\scriptsize \textcolor{green!60!black}{\rule{3pt}{5pt}}} & 53.24 {\scriptsize \textcolor{green!60!black}{\rule{1.72pt}{5pt}}} & 56.95 {\scriptsize \textcolor{green!60!black}{\rule{0.61pt}{5pt}}} & 78.45 {\scriptsize \textcolor{green!60!black}{\rule{6.23pt}{5pt}}} & 76.40 {\scriptsize \textcolor{red}{\rule{5.63pt}{5pt}}} & 80.61 {\scriptsize \textcolor{red}{\rule{0.36pt}{5pt}}} & 54.96 {\scriptsize \textcolor{red}{\rule{0.06pt}{5pt}}} & 37.84 {\scriptsize \textcolor{red}{\rule{6.49pt}{5pt}}} & 96.43 {\scriptsize \textcolor{green!60!black}{\rule{1.93pt}{5pt}}} \\

SFT(WS) & 30.75 {\scriptsize \textcolor{green!60!black}{\rule{1.72pt}{5pt}}} & 48.72 {\scriptsize \textcolor{red}{\rule{2.8pt}{5pt}}} & 52.39 {\scriptsize \textcolor{red}{\rule{3.95pt}{5pt}}} & 76.60 {\scriptsize \textcolor{green!60!black}{\rule{4.39pt}{5pt}}} & 75.50 {\scriptsize \textcolor{red}{\rule{6.53pt}{5pt}}} & 76.58 {\scriptsize \textcolor{red}{\rule{1.02pt}{5pt}}} & 54.86 {\scriptsize \textcolor{red}{\rule{0.16pt}{5pt}}} & 36.01 {\scriptsize \textcolor{red}{\rule{8.32pt}{5pt}}} & 94.59 {\scriptsize \textcolor{green!60!black}{\rule{0.09pt}{5pt}}}\\

SFT(R) & 30.00 {\scriptsize \textcolor{green!60!black}{\rule{0.97pt}{5pt}}} & 52.80 {\scriptsize \textcolor{green!60!black}{\rule{1.28pt}{5pt}}} & 56.02 {\scriptsize \textcolor{red}{\rule{0.32pt}{5pt}}} & 77.42 {\scriptsize \textcolor{green!60!black}{\rule{5.2pt}{5pt}}} & 77.15 {\scriptsize \textcolor{green!60!black}{\rule{4.88pt}{5pt}}} & 76.83 {\scriptsize \textcolor{red}{\rule{4.14pt}{5pt}}} & 56.40 {\scriptsize \textcolor{green!60!black}{\rule{1.38pt}{5pt}}} & 41.06 {\scriptsize \textcolor{red}{\rule{3.27pt}{5pt}}} & 88.40 {\scriptsize \textcolor{red}{\rule{6.1pt}{5pt}}}\\

SFT(CS)+DPO(Dial.)& 28.81 {\scriptsize \textcolor{red}{\rule{0.22pt}{5pt}}} & 45.61 {\scriptsize \textcolor{red}{\rule{5.91pt}{5pt}}} & 57.22 {\scriptsize \textcolor{green!60!black}{\rule{0.88pt}{5pt}}} & 64.09 {\scriptsize \textcolor{red}{\rule{8.13pt}{5pt}}} & 80.24 {\scriptsize \textcolor{red}{\rule{1.79pt}{5pt}}} & 81.46 {\scriptsize \textcolor{green!60!black}{\rule{0.49pt}{5pt}}} & 48.06 {\scriptsize \textcolor{red}{\rule{6.96pt}{5pt}}} & 45.47 {\scriptsize \textcolor{red}{\rule{1.14pt}{5pt}}} & 84.22 {\scriptsize \textcolor{red}{\rule{10.28pt}{5pt}}}\\

SFT(CS)+DPO(Turn)& 28.75 {\scriptsize \textcolor{red}{\rule{0.28pt}{5pt}}} & 48.09 {\scriptsize \textcolor{red}{\rule{3.43pt}{5pt}}} & 56.15 {\scriptsize \textcolor{red}{\rule{0.19pt}{5pt}}} & 67.88 {\scriptsize \textcolor{red}{\rule{4.34pt}{5pt}}} & 81.93 {\scriptsize \textcolor{red}{\rule{0.1pt}{5pt}}} {\scriptsize \textcolor{green!60!black}{\rule{2.53pt}{5pt}}} & 83.05 {\scriptsize \textcolor{green!60!black}{\rule{2.08pt}{5pt}}} & 52.87 {\scriptsize \textcolor{red}{\rule{2.15pt}{5pt}}} & 48.26 {\scriptsize \textcolor{green!60!black}{\rule{3.93pt}{5pt}}} & 84.92 {\scriptsize \textcolor{red}{\rule{9.58pt}{5pt}}}\\

\textbf{Qwen-2-7B} &&&&&&&&\\

Base & 07.40 & 37.21  & 52.99  & 65.51 & 71.42  & 61.22  & 52.25  & 32.17 & 65.82 \\

SFT(CS) & 06.94 {\scriptsize \textcolor{red}{\rule{0.46pt}{5pt}}} & 37.66 {\scriptsize \textcolor{green!60!black}{\rule{0.45pt}{5pt}}}  & 51.64 {\scriptsize \textcolor{red}{\rule{1.35pt}{5pt}}} & 65.43 {\scriptsize \textcolor{red}{\rule{0.08pt}{5pt}}} & 55.61 {\scriptsize \textcolor{red}{\rule{15.81pt}{5pt}}} &  61.10 {\scriptsize \textcolor{red}{\rule{0.12pt}{5pt}}} & 51.64 {\scriptsize \textcolor{red}{\rule{0.61pt}{5pt}}} & 37.10 {\scriptsize \textcolor{green!60!black}{\rule{4.93pt}{5pt}}} & 75.33 {\scriptsize \textcolor{green!60!black}{\rule{9.51pt}{5pt}}} \\

SFT(CS)+DPO(Dial.)& 6.70 {\scriptsize \textcolor{red}{\rule{0.7pt}{5pt}}} & 35.69 {\scriptsize \textcolor{red}{\rule{1.52pt}{5pt}}}  & 51.93 {\scriptsize \textcolor{red}{\rule{1.06pt}{5pt}}} & 62.90 {\scriptsize \textcolor{red}{\rule{2.61pt}{5pt}}} & 57.63 {\scriptsize \textcolor{red}{\rule{13.79pt}{5pt}}} & 64.39 {\scriptsize \textcolor{green!60!black}{\rule{3.17pt}{5pt}}} & 49.84 {\scriptsize \textcolor{red}{\rule{2.41pt}{5pt}}}  & 32.99 {\scriptsize \textcolor{green!60!black}{\rule{0.82pt}{5pt}}} & 79.08 {\scriptsize \textcolor{green!60!black}{\rule{13.26pt}{5pt}}} \\

SFT(CS)+DPO(Turn)& 6.80 {\scriptsize \textcolor{red}{\rule{0.6pt}{5pt}}} & 37.72 {\scriptsize \textcolor{green!60!black}{\rule{0.51pt}{5pt}}} & 51.74 {\scriptsize \textcolor{red}{\rule{1.25pt}{5pt}}} & 60.62 {\scriptsize \textcolor{red}{\rule{4.89pt}{5pt}}}  & 61.84 {\scriptsize \textcolor{red}{\rule{9.58pt}{5pt}}} & 64.88 {\scriptsize \textcolor{green!60!black}{\rule{3.66pt}{5pt}}} & 48.57 {\scriptsize \textcolor{red}{\rule{3.68pt}{5pt}}}  & 34.57 {\scriptsize \textcolor{green!60!black}{\rule{2.4pt}{5pt}}}  & 84.31 {\scriptsize \textcolor{green!60!black}{\rule{18.49pt}{5pt}}} \\

\bottomrule
\end{tabular}
}
\caption{\textbf{Performance on Executive and Socio-Emotional Tasks.} SimpleToM (AJ) and (ToM) are grouped based on the taxonomy in \citet{momente2025triangulating}.}
\label{tab:reference-based-results-detailed-1}
\end{table*}

\begin{table*}[ht]
\centering
\resizebox{\textwidth}{!}{%
\begin{tabular}{ll|ll|l}
\hline
 &
  \multicolumn{1}{c|}{\textbf{Formal}} &
  \multicolumn{2}{c|}{\textbf{General}} &
  \multicolumn{1}{c}{\textbf{Instruction-following}} \\
 &
  \multicolumn{1}{c|}{GLUE Diagnostics} &
  \multicolumn{1}{c}{MMLU-Pro} &
  \multicolumn{1}{c|}{BBH} &
  \multicolumn{1}{c}{IFEval} \\ \hline
\textbf{Llama-3.1-8B} &
   &
   &
   &
   \\
Base &
  38.06 &
  43.35 &
  40.37 &
  76.88 \\
SFT (CS) &
  40.23 {\scriptsize \textcolor{green!60!black}{\rule{2.17pt}{5pt}}} &
  13.16 {\scriptsize \textcolor{red}{\rule{29.75pt}{5pt}}} &
  46.75 {\scriptsize \textcolor{green!60!black}{\rule{6.38pt}{5pt}}} &
  67.25 {\scriptsize \textcolor{red}{\rule{9.63pt}{5pt}}} \\
SFT (WS) &
  30.74 {\scriptsize \textcolor{red}{\rule{7.32pt}{5pt}}} &
  01.70 {\scriptsize \textcolor{red}{\rule{41.65pt}{5pt}}} &
  45.52 {\scriptsize \textcolor{green!60!black}{\rule{5.15pt}{5pt}}} &
  61.40 {\scriptsize \textcolor{red}{\rule{15.48pt}{5pt}}} \\
SFT (R) &
  38.62 {\scriptsize \textcolor{green!60!black}{\rule{0.56pt}{5pt}}} &
  31.99 {\scriptsize \textcolor{red}{\rule{11.36pt}{5pt}}} &
  45.66 {\scriptsize \textcolor{green!60!black}{\rule{5.29pt}{5pt}}} &
  68.76 {\scriptsize \textcolor{red}{\rule{8.12pt}{5pt}}} \\
SFT (CS) + DPO (Dial.) &
  36.20 {\scriptsize \textcolor{red}{\rule{1.86pt}{5pt}}} &
  09.28 {\scriptsize \textcolor{red}{\rule{34.07pt}{5pt}}} &
  43.86 {\scriptsize \textcolor{green!60!black}{\rule{3.49pt}{5pt}}} &
  68.39 {\scriptsize \textcolor{red}{\rule{8.49pt}{5pt}}} \\
SFT (CS) + DPO (Turn.) &
  36.07 {\scriptsize \textcolor{red}{\rule{0.86pt}{1.99pt}}} &
  11.13 {\scriptsize \textcolor{red}{\rule{32.22pt}{5pt}}} &
  46.80 {\scriptsize \textcolor{green!60!black}{\rule{6.43pt}{5pt}}} &
  70.76 {\scriptsize \textcolor{green!60!black}{\rule{6.12pt}{5pt}}} \\
GRPO &
  38.68 {\scriptsize \textcolor{green!60!black}{\rule{0.62pt}{5pt}}} &
  43.73 {\scriptsize \textcolor{green!60!black}{\rule{0.38pt}{5pt}}} &
  39.31 {\scriptsize \textcolor{red}{\rule{1.06pt}{5pt}}} &
  76.97 {\scriptsize \textcolor{green!60!black}{\rule{0.09pt}{5pt}}} \\
SFT (CS) + GRPO &
  37.31 {\scriptsize \textcolor{red}{\rule{0.75pt}{5pt}}} &
  41.55 {\scriptsize \textcolor{red}{\rule{1.8pt}{5pt}}} &
  44.09 {\scriptsize \textcolor{green!60!black}{\rule{3.72pt}{5pt}}} &
  75.77 {\scriptsize \textcolor{red}{\rule{1.11pt}{5pt}}} \\
 &
   &
   &
   &
   \\
\textbf{Llama-3.1-70B} &
   &
   &
   &
   \\
Base &
  46.16 &
  60.37 &
  60.74 &
  85.16 \\
SFT(CS) &
  47.72 {\scriptsize \textcolor{green!60!black}{\rule{1.56pt}{5pt}}} &
  25.90 {\scriptsize \textcolor{red}{\rule{34.47pt}{5pt}}} &
  63.91 {\scriptsize \textcolor{green!60!black}{\rule{3.17pt}{5pt}}} &
  79.38 {\scriptsize \textcolor{red}{\rule{5.78pt}{5pt}}} \\
SFT (WS) &
  45.86 {\scriptsize \textcolor{red}{\rule{0.3pt}{5pt}}} &
  25.03 {\scriptsize \textcolor{red}{\rule{35.34pt}{5pt}}} &
  63.58 {\scriptsize \textcolor{green!60!black}{\rule{2.84pt}{5pt}}} &
  75.10 {\scriptsize \textcolor{red}{\rule{10.06pt}{5pt}}} \\
SFT (R) &
  46.51 {\scriptsize \textcolor{green!60!black}{\rule{0.35pt}{5pt}}} &
  18.30 {\scriptsize \textcolor{red}{\rule{42.07pt}{5pt}}} &
  65.63 {\scriptsize \textcolor{green!60!black}{\rule{4.89pt}{5pt}}} &
  79.68 {\scriptsize \textcolor{red}{\rule{5.48pt}{5pt}}} \\
SFT(CS)+DPO(Dial.) &
  37.73 {\scriptsize \textcolor{red}{\rule{8.43pt}{5pt}}} &
  38.34 {\scriptsize \textcolor{red}{\rule{22.03pt}{5pt}}} &
  39.21 {\scriptsize \textcolor{red}{\rule{21.53pt}{5pt}}} &
  82.26 {\scriptsize \textcolor{red}{\rule{2.9pt}{5pt}}} \\
SFT(CS)+DPO(Turn) &
  39.23 {\scriptsize \textcolor{red}{\rule{6.93pt}{5pt}}} &
  36.02 {\scriptsize \textcolor{red}{\rule{24.35pt}{5pt}}} &
  53.69 {\scriptsize \textcolor{red}{\rule{7.05pt}{5pt}}} &
  85.68 {\scriptsize \textcolor{green!60!black}{\rule{0.52pt}{5pt}}} \\
 &
   &
   &
   &
   \\
\textbf{Qwen-2-7B} & 
   & 
   & 
   & 
   \\
Base & 59.68
   & 00.42
   & 33.24
   & 59.16
   \\
SFT(CS) & 52.98 {\scriptsize \textcolor{red}{\rule{6.7pt}{5pt}}}
   & 00.18 {\scriptsize \textcolor{red}{\rule{0.24pt}{5pt}}}
   & 42.53 {\scriptsize \textcolor{green!60!black}{\rule{9.29pt}{5pt}}}
   & 53.87 {\scriptsize \textcolor{red}{\rule{5.29pt}{5pt}}}
   \\
SFT(CS)+DPO(Dial.) & 53.74 {\scriptsize \textcolor{red}{\rule{5.94pt}{5pt}}}
   & 00.71 {\scriptsize \textcolor{green!60!black}{\rule{0.29pt}{5pt}}}
   & 29.99 {\scriptsize \textcolor{red}{\rule{3.25pt}{5pt}}}
   & 54.20 {\scriptsize \textcolor{red}{\rule{5.06pt}{5pt}}}
   \\
SFT(CS)+DPO(Turn) & 48.34 {\scriptsize \textcolor{red}{\rule{11.34pt}{5pt}}}
   & 00.07 {\scriptsize \textcolor{red}{\rule{0.35pt}{5pt}}} 
   & 27.34 {\scriptsize \textcolor{red}{\rule{5.9pt}{5pt}}}
   & 54.99 {\scriptsize \textcolor{red}{\rule{4.27pt}{5pt}}}
   \\ \hline
\end{tabular}%
}
\caption{\textbf{Model performance on formal, general and instruction-following capabilities}, as measured by GLUE Diagnostics \cite{wang2018glue}, MMLU-Pro and BBH \cite{mmlupro2024,suzgun-etal-2023-challenging}, IFEval \cite{ifeval}.}
\label{tab:reference-based-results-detailed-2}
\end{table*}

\section{Qualitative Discussion}
\label{app:error_analysis}

\begin{figure*}
    \centering
    \includegraphics[width=\linewidth]{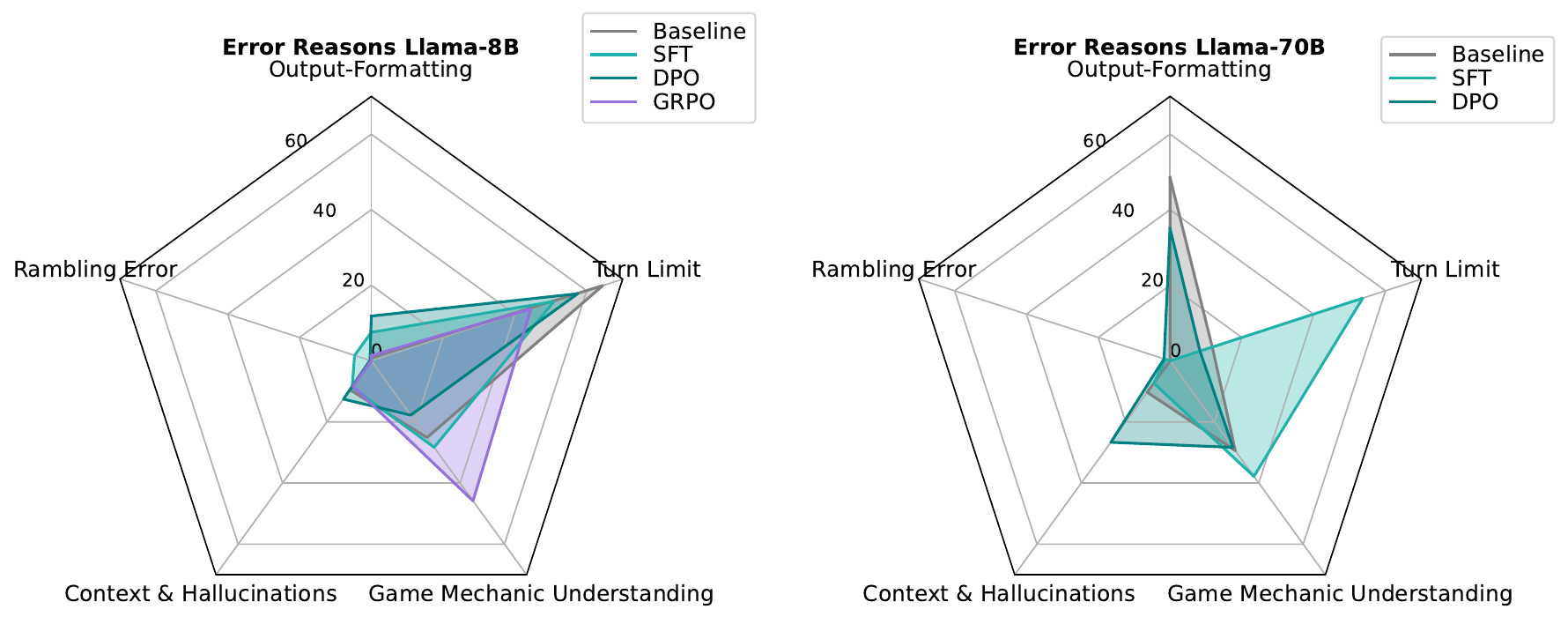}
    \caption{Relative distribution of error categories.}
    \label{fig:radar_chart_error_analysis}
\end{figure*}

While we know that the absolute number of aborted episodes goes down from the baseline to SFT to GRPO, the distribution of reasons for those aborted episodes might change. For out-of-domain games, we investigated these reasons. Five overarching error categories were manually grouped together; the relative distributions of errors between the different models are depicted in Figure \ref{fig:radar_chart_error_analysis}. The main problem for the base Llama-3.1-8B is that of exceeding the turn limit (e.g. for the Map Navigation game, continuing to loop between already visited rooms), whereas for Llama-3.1-70B problems are mostly with the answer format. This verbose behaviour is reduced by all the types of training performed. The best performing Llama-3.1-8B version is the GRPO one, achieving the lowest absolute number in terms of aborted episodes, with fewer turn limit errors and more game mechanic understanding ones in proportion. Both for Llama-3.1-8B and Llama-3.1-70B, DPO shows the highest proportion of hallucination and context-related errors, a possible signal of overfitting to the training data.

\begin{figure*}
    \centering
    \includegraphics[width=.98\linewidth]{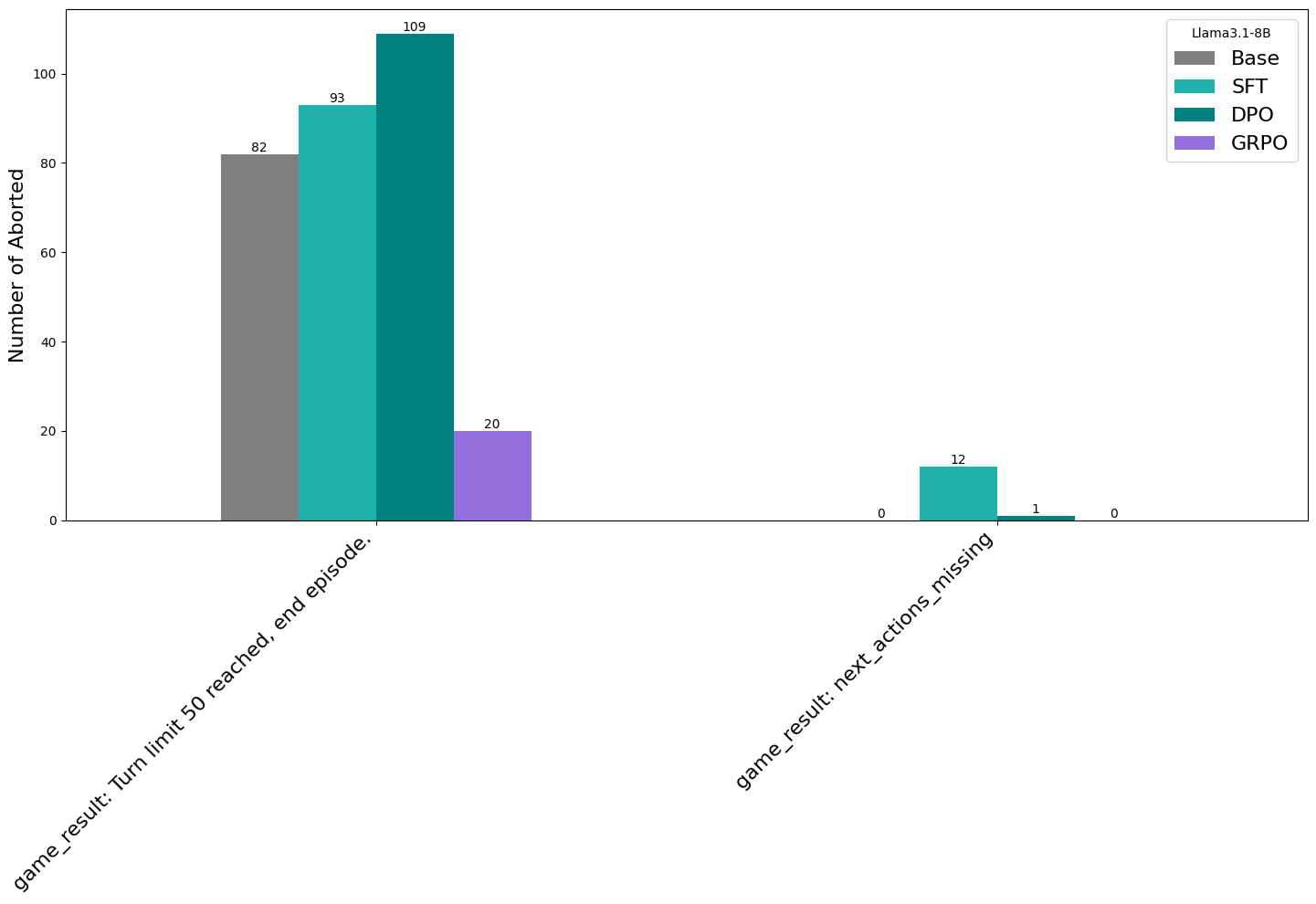}
    \caption{Number of aborted interactions per abortion reason in \textbf{Adventuregame.}}
    \label{fig:appendix_analysis_adv}
\end{figure*}

\begin{figure*}
    \centering
    \includegraphics[width=.98\linewidth]{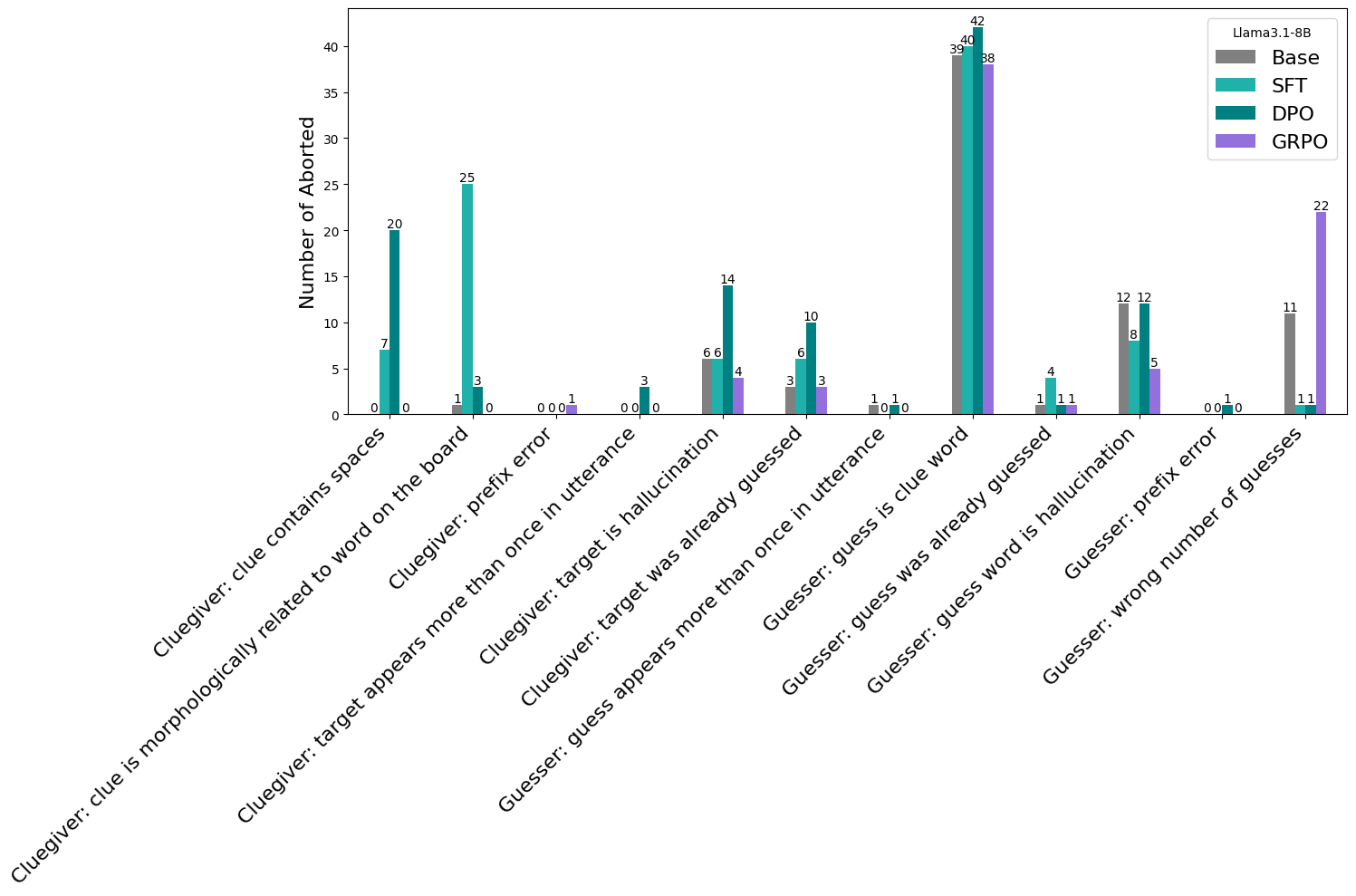}
    \caption{Number of aborted interactions per abortion reason in \textbf{Codenames}.}
    \label{fig:appendix_analysis_codenames}
\end{figure*}

In Section \ref{sec:results}, we show how Llama-3.1-8B trained with GRPO data is able to generalise to out-of-domain games. One of the main reasons for this is the reduced number of aborted games due to exceeding the game's turn limit (Fig. \ref{fig:radar_chart_error_analysis}). Fig. \ref{fig:appendix_analysis_adv} reports the absolute number of aborted episodes per possible aborted reasons in the Adventure Game for Llama-3.1-8B. The GRPO trained version drastically reduces the number of overall errors in the game, with around 1/4 of the original aborted episodes due to reaching the turn limit, and not reproducing the rambling errors of the SFT version (\lq\lq next\_action\_missing''). Adventure Game, where the GRPO's higher percentage of played games (+19.62) is coupled with a higher quality score (+11.28), is a good example of the stability of these out-of-domain gains by GRPO. For Codenames, Fig. \ref{fig:appendix_analysis_codenames} reports the absolute number of aborted episodes per possible aborted reasons for Llama-3.1-8B. We observe for GRPO a reduced number of hallucinations (\lq\lq Target is hallucination'' and \lq\lq Guess word is hallucination'') compared to the base and SFT, while not decreasing in most cases compared to the baseline. A notable exception is the \lq\lq Wrong number of guesses'', where the GRPO model seems not to respect the number of guesses per turn required by the game. Finally, we report a Codenames episode played by all the base Llama-3.1-8B, SFT, and GRPO. As shown in Fig. \ref{tab:game-continuations}, the base model fails due to the common error of guessing the clue word; the SFT does not encounter errors but reveals the killer word, losing the game. The GRPO, instead, is able to play the game successfully.

\footnotesize
\setcounter{utterance}{1}
\renewcommand{\arraystretch}{1.2}
\setlength{\tabcolsep}{3pt}

\onecolumn  

\begin{longtable}{c|p{0.31\textwidth}|p{0.31\textwidth}|p{0.31\textwidth}|}

\caption{Codenames sample by the Base model, the SFT and GRPO trained ones.}
\label{tab:game-continuations} \\

\theutterance &
\multicolumn{3}{p{0.95\textwidth}|}{
\colorbox[rgb]{0.9,0.9,0.9}{%
\begin{minipage}{0.93\textwidth}
\texttt{\tiny [A$\langle$GM]}\\
\texttt{Provide one single-word clue that relates to some of your team's words from the list below. You can choose to how many and to which words your clue relates to.}\\
\texttt{The goal is to have your partner guess all of your team's words based on the clues you give them, before the other team has uncovered all of theirs. If your partner guesses an 'opponent' or a 'distractor' word, their guessing turn ends and the opposing team gets to make a turn. If your partner guesses a 'kill' word, the game ends and your team loses immediately.}\\
\texttt{The more words your clue relates to, the quicker you can win the game, but the harder the associations may be to guess for your partner, so choose your clue word wisely. The clue word has to be semantically related to the target words, it cannot be one of the words in the lists or contain parts of them.}\\
\texttt{}\\
\texttt{Always give your single-word clue and your comma-separated list of related target words in the following format and make your answers as short as possible, never include any other text than is required in this form:}\\
\texttt{}\\
\texttt{CLUE: <WORD>}\\
\texttt{TARGETS: <TARGETS>}\\
\texttt{}\\
\texttt{Your team words are: gospel, shears, thunder, fog, bluegrass, attic, entrance, basement, hose.}\\
\texttt{Your opponent's team words are: cotton, head, buttocks, running, back, wool, swimming, neck.}\\
\texttt{Distractor words are: seasoning, brazil, oregano, china, portugal, truck, atv.}\\
\texttt{The kill words are: silver.}
\end{minipage}}} \\
\hline

\multicolumn{1}{c|}{} & \multicolumn{1}{c|}{\textbf{Base}} & \multicolumn{1}{c|}{\textbf{SFT}} & \multicolumn{1}{c|}{\textbf{GRPO}} \\
\hline
\endfirsthead

\multicolumn{4}{c}{Table \thetable{}} \\
\hline
\multicolumn{1}{c|}{\#} & \multicolumn{1}{c|}{\textbf{Base}} & \multicolumn{1}{c|}{\textbf{SFT}} & \multicolumn{1}{c|}{\textbf{GRPO}} \\
\hline
\endhead

\hline
\multicolumn{4}{r}{\textit{Continued on next page}} \\
\endfoot

\hline
\endlastfoot

\stepcounter{utterance}\theutterance &
\colorbox[rgb]{0.8,1,0.9}{%
\begin{minipage}{0.3\textwidth}
\texttt{\tiny [A$\rangle$GM]}\\
\texttt{CLUE: farm}\\
\texttt{TARGETS: bluegrass, shears, gospel}
\end{minipage}} &

\colorbox[rgb]{0.8,1,0.9}{%
\begin{minipage}{0.3\textwidth}
\texttt{\tiny [A$\rangle$GM]}\\
\texttt{CLUE: roof}\\
\texttt{TARGETS: attic, entrance, basement}
\end{minipage}} &

\colorbox[rgb]{0.8,1,0.9}{%
\begin{minipage}{0.3\textwidth}
\texttt{\tiny [A$\rangle$GM]}\\
\texttt{CLUE: farm}\\
\texttt{TARGETS: bluegrass, shears, gospel}
\end{minipage}} \\
\hline

\stepcounter{utterance}\theutterance &
\colorbox[rgb]{0.9,0.9,0.9}{%
\begin{minipage}{0.3\textwidth}
\texttt{\tiny [GM$\rangle$B]}\\
\texttt{Provide a comma-separated list of up to 2 words from the following list that best relate or are most closely associated with the word '\textbf{farm}'.}\\
\texttt{Always start your list of guess(es) with 'GUESS: ' and do not include any other text in your answer.}\\
\texttt{}\\
\texttt{attic, running, neck, bluegrass, atv, brazil, portugal, seasoning, entrance, truck, china, buttocks, head, oregano, fog, swimming, wool, gospel, cotton, basement, hose, shears, back, thunder, silver}
\end{minipage}} &

\colorbox[rgb]{0.9,0.9,0.9}{%
\begin{minipage}{0.3\textwidth}
\texttt{\tiny [GM$\rangle$B]}\\
\texttt{Provide a comma-separated list of up to 3 words from the following list that best relate or are most closely associated with the word 'roof'. Always start your list of guess(es) with 'GUESS: ' and do not include any other text in your answer.}\\
\texttt{}\\
\texttt{attic, running, neck, bluegrass, atv, brazil, portugal, seasoning, entrance, truck, china, buttocks, head, oregano, fog, swimming, wool, gospel, cotton, basement, hose, shears, back, thunder, silver}
\end{minipage}} &

\colorbox[rgb]{0.9,0.9,0.9}{%
\begin{minipage}{0.3\textwidth}
\texttt{\tiny [GM$\rangle$B]}\\
\texttt{Provide a comma-separated list of up to 3 words from the following list that best relate or are most closely associated with the word 'farm'. Always start your list of guess(es) with 'GUESS: ' and do not include any other text in your answer.}\\
\texttt{}\\
\texttt{attic, running, neck, bluegrass, atv, brazil, portugal, seasoning, entrance, truck, china, buttocks, head, oregano, fog, swimming, wool, gospel, cotton, basement, hose, shears, back, thunder, silver}
\end{minipage}} \\
\hline

\stepcounter{utterance}\theutterance &
\colorbox[rgb]{1,0.85,0.72}{%
\begin{minipage}{0.3\textwidth}
\texttt{\tiny [GM$\langle$B]}\\
\texttt{GUESS: wool, shears, cotton}
\end{minipage}} &

\colorbox[rgb]{1,0.85,0.72}{%
\begin{minipage}{0.3\textwidth}
\texttt{\tiny [GM$\langle$B]}\\
\texttt{GUESS: attic, head, back}
\end{minipage}} &

\colorbox[rgb]{1,0.85,0.72}{%
\begin{minipage}{0.3\textwidth}
\texttt{\tiny [GM$\langle$B]}\\
\texttt{GUESS: bluegrass, cotton, shears}
\end{minipage}} \\
\hline

\stepcounter{utterance}\theutterance &
\colorbox[rgb]{0.95,0.95,0.95}{%
\begin{minipage}{0.3\textwidth}
\texttt{\tiny [A$\langle$GM]}\\
\texttt{The word wool was guessed but is an opponent word. Your teammate's turn ended there.
Now provide another clue relating to some of your remaining team words and a list of the related target words. Remember to start your clue with 'CLUE: ', put a new line, and start your comma-separated list of target words with 'TARGETS: '. Notice: some words have been removed from the lists compared to previous requests.\\
\texttt{}\\
Your remaining team words are: gospel, shears, thunder, fog, bluegrass, attic, entrance, basement, hose. \\
Remaining words for your opponent are: cotton, head, buttocks, running, back, swimming, neck. \\
Remaining distractor words are: seasoning, brazil, oregano, china, portugal, truck, atv.
Remaining kill words are: silver.}
\end{minipage}} &

\colorbox[rgb]{0.95,0.95,0.95}{%
\begin{minipage}{0.3\textwidth}
\texttt{\tiny [A$\langle$GM]}\\
\texttt{The word attic was guessed correctly. The word head was guessed but is an opponent word. Your teammate's turn ended there.
Now provide another clue relating to some of your remaining team words and a list of the related target words. Remember to start your clue with 'CLUE: ', put a new line, and start your comma-separated list of target words with 'TARGETS: '. Notice: some words have been removed from the lists compared to previous requests.\\
\texttt{}\\
Your remaining team words are: gospel, shears, thunder, fog, bluegrass, entrance, basement, hose.
Remaining words for your opponent are: cotton, buttocks, running, back, wool, swimming, neck.
Remaining distractor words are: seasoning, brazil, oregano, china, portugal, truck, atv.
Remaining kill words are: silver.}
\end{minipage}}&

\colorbox[rgb]{0.95,0.95,0.95}{%
\begin{minipage}{0.3\textwidth}
\texttt{\tiny [A$\langle$GM]}\\
\texttt{The word bluegrass was guessed correctly. The word cotton was guessed but is an opponent word. Your teammate's turn ended there.
Now provide another clue relating to some of your remaining team words and a list of the related target words. Remember to start your clue with 'CLUE: ', put a new line, and start your comma-separated list of target words with 'TARGETS: '. Notice: some words have been removed from the lists compared to previous requests.\\
\texttt{}\\
Your remaining team words are: gospel, shears, thunder, fog, attic, entrance, basement, hose.
Remaining words for your opponent are: head, buttocks, running, back, wool, swimming, neck.
Remaining distractor words are: seasoning, brazil, oregano, china, portugal, truck, atv.
Remaining kill words are: silver.}
\end{minipage}} \\ \hline

\stepcounter{utterance}\theutterance &

\colorbox[rgb]{0.8,1,0.9}{%
\begin{minipage}{0.3\textwidth}
\texttt{\tiny [A$\rangle$GM]}\\
\texttt{CLUE: storm}\\
\texttt{TARGETS: thunder, fog, hose}
\end{minipage}} &

\colorbox[rgb]{0.8,1,0.9}{%
\begin{minipage}{0.3\textwidth}
\texttt{\tiny [A$\rangle$GM]}\\
\texttt{CLUE: storm}\\
\texttt{TARGETS: thunder, fog, hose}
\end{minipage}}   &

\colorbox[rgb]{0.8,1,0.9}{%
\begin{minipage}{0.3\textwidth}
\texttt{\tiny [A$\rangle$GM]}\\
\texttt{CLUE: storm}\\
\texttt{TARGETS: thunder, fog}
\end{minipage}}  \\
\hline

\stepcounter{utterance}\theutterance &
\colorbox[rgb]{0.95,0.95,0.95}{%
\begin{minipage}{0.3\textwidth}
\texttt{\tiny [GM$\rangle$B]}\\
\texttt{The word wool was guessed but is an opponent word. Your turn ended there.
Now provide another comma-separated list of at least 1 and up to 3 words from the following list of words that best relate or are most closely associated with the word 'storm'. Remember to start your answer with 'GUESS: '. Notice: some words have been removed from the list compared to previous requests.\\
\texttt{}\\
attic, running, neck, bluegrass, atv, brazil, portugal, seasoning, entrance, truck, china, buttocks, head, oregano, fog, gospel, cotton, basement, hose, shears, back, thunder, silver}
\end{minipage}} &

\colorbox[rgb]{0.95,0.95,0.95}{%
\begin{minipage}{0.3\textwidth}
\texttt{\tiny [GM$\rangle$B]}\\
\texttt{The word attic was guessed correctly. The word head was guessed but is an opponent word. Your turn ended there.
Now provide another comma-separated list of at least 1 and up to 3 words from the following list of words that best relate or are most closely associated with the word 'storm'. Remember to start your answer with 'GUESS: '. Notice: some words have been removed from the list compared to previous requests.\\
\texttt{}\\
neck, bluegrass, atv, brazil, portugal, seasoning, entrance, truck, china, buttocks, oregano, fog, swimming, wool, gospel, cotton, basement, hose, shears, back, thunder, silver}
\end{minipage}} 
&

\colorbox[rgb]{0.95,0.95,0.95}{%
\begin{minipage}{0.3\textwidth}
\texttt{\tiny [GM$\rangle$B]}\\
\texttt{The word bluegrass was guessed correctly. The word cotton was guessed but is an opponent word. Your turn ended there.
Now provide another comma-separated list of at least 1 and up to 2 words from the following list of words that best relate or are most closely associated with the word 'storm'. Remember to start your answer with 'GUESS: '. Notice: some words have been removed from the list compared to previous requests.\\
\texttt{}\\
attic, running, neck, atv, brazil, portugal, seasoning, entrance, truck, china, buttocks, head, oregano, fog, swimming, wool, gospel, basement, hose, shears, thunder, silver}
\end{minipage}} \\
\hline

\stepcounter{utterance}\theutterance &
\colorbox[rgb]{1,0.85,0.72}{%
\begin{minipage}{0.3\textwidth}
\texttt{\tiny [GM$\langle$B]}\\
\texttt{GUESS: fog, thunder, basement}
\end{minipage}} &
\colorbox[rgb]{1,0.85,0.72}{%
\begin{minipage}{0.3\textwidth}
\texttt{\tiny [GM$\langle$B]}\\
\texttt{GUESS: fog, thunder, back}
\end{minipage}} &
\colorbox[rgb]{1,0.85,0.72}{%
\begin{minipage}{0.3\textwidth}
\texttt{\tiny [GM$\langle$B]}\\
\texttt{GUESS: fog, thunder}
\end{minipage}}  \\
\hline

\stepcounter{utterance}\theutterance &
\colorbox[rgb]{0.95,0.95,0.95}{%
\begin{minipage}{0.3\textwidth}
\texttt{\tiny [A$\langle$GM]}\\
\texttt{The words fog, thunder, basement were guessed correctly. Your teammate's turn ended there. 
Now provide another clue relating to some of your remaining team words and a list of the related target words. Remember to start your clue with 'CLUE: ', put a new line, and start your comma-separated list of target words with 'TARGETS: '. Notice: some words have been removed from the lists compared to previous requests.\\
\texttt{}\\
Your remaining team words are: gospel, shears, bluegrass, attic, entrance, hose.
Remaining words for your opponent are: cotton, head, buttocks, running, back, neck.
Remaining distractor words are: seasoning, brazil, oregano, china, portugal, truck, atv.
Remaining kill words are: silver.}
\end{minipage}}
&
\colorbox[rgb]{0.95,0.95,0.95}{%
\begin{minipage}{0.3\textwidth}
\texttt{\tiny [A$\langle$GM]}\\
\texttt{The words fog, thunder were guessed correctly. The word back was guessed but is an opponent word. Your teammate's turn ended there.
Now provide another clue relating to some of your remaining team words and a list of the related target words. Remember to start your clue with 'CLUE: ', put a new line, and start your comma-separated list of target words with 'TARGETS: '. Notice: some words have been removed from the lists compared to previous requests.\\
\texttt{}\\
Your remaining team words are: gospel, shears, bluegrass, entrance, basement, hose.
Remaining words for your opponent are: cotton, buttocks, wool, swimming, neck.
Remaining distractor words are: seasoning, brazil, oregano, china, portugal, truck, atv.
Remaining kill words are: silver.}
\end{minipage}}
&
\colorbox[rgb]{0.95,0.95,0.95}{%
\begin{minipage}{0.3\textwidth}
\texttt{\tiny [A$\langle$GM]}\\
\texttt{The words fog, thunder were guessed correctly. Your teammate's turn ended there.
Now provide another clue relating to some of your remaining team words and a list of the related target words. Remember to start your clue with 'CLUE: ', put a new line, and start your comma-separated list of target words with 'TARGETS: '. Notice: some words have been removed from the lists compared to previous requests.\\
\texttt{}\\
Your remaining team words are: gospel, shears, attic, entrance, basement, hose.
Remaining words for your opponent are: head, buttocks, running, wool, swimming, neck.
Remaining distractor words are: seasoning, brazil, oregano, china, portugal, truck, atv.
Remaining kill words are: silver.}
\end{minipage}} \\
\hline

\stepcounter{utterance}\theutterance &
\colorbox[rgb]{0.8,1,0.9}{%
\begin{minipage}{0.3\textwidth}
\texttt{\tiny [A$\rangle$GM]}\\
\texttt{CLUE: music}\\
\texttt{TARGETS: gospel, bluegrass}
\end{minipage}}
&
\colorbox[rgb]{0.8,1,0.9}{%
\begin{minipage}{0.3\textwidth}
\texttt{\tiny [A$\rangle$GM]}\\
\texttt{CLUE: home}\\
\texttt{TARGETS: entrance, basement, hose}
\end{minipage}}
&
\colorbox[rgb]{0.8,1,0.9}{%
\begin{minipage}{0.3\textwidth}
\texttt{\tiny [A$\rangle$GM]}\\
\texttt{CLUE: home}\\
\texttt{TARGETS: attic, entrance, basement}
\end{minipage}}\\
\hline

\stepcounter{utterance}\theutterance &
\colorbox[rgb]{0.95,0.95,0.95}{%
\begin{minipage}{0.3\textwidth}
\texttt{\tiny [GM$\rangle$B]}\\
\texttt{The words fog, thunder, basement were guessed correctly. Your turn ended there.
Now provide another comma-separated list of at least 1 and up to 2 words from the following list of words that best relate or are most closely associated with the word 'music'. Remember to start your answer with 'GUESS: '. Notice: some words have been removed from the list compared to previous requests.\\
\texttt{}\\
attic, running, neck, bluegrass, atv, brazil, portugal, seasoning, entrance, truck, china, buttocks, head, oregano, gospel, hose, shears, back, silver}
\end{minipage}}
&
\colorbox[rgb]{0.95,0.95,0.95}{%
\begin{minipage}{0.3\textwidth}
\texttt{\tiny [GM$\rangle$B]}\\
\texttt{The words fog, thunder were guessed correctly. The word back was guessed but is an opponent word. Your turn ended there.
Now provide another comma-separated list of at least 1 and up to 3 words from the following list of words that best relate or are most closely associated with the word 'home'. Remember to start your answer with 'GUESS: '. Notice: some words have been removed from the list compared to previous requests.\\
\texttt{}\\
neck, bluegrass, atv, brazil, portugal, seasoning, entrance, truck, china, buttocks, oregano, swimming, gospel, cotton, basement, hose, shears, silver}
\end{minipage}}
&
\colorbox[rgb]{0.95,0.95,0.95}{%
\begin{minipage}{0.3\textwidth}
\texttt{\tiny [GM$\rangle$B]}\\
\texttt{The words fog, thunder were guessed correctly. Your turn ended there.
Now provide another comma-separated list of at least 1 and up to 3 words from the following list of words that best relate or are most closely associated with the word 'home'. Remember to start your answer with 'GUESS: '. Notice: some words have been removed from the list compared to previous requests.\\
\texttt{}\\
attic, running, neck, atv, brazil, portugal, seasoning, entrance, truck, china, buttocks, oregano, swimming, wool, gospel, basement, hose, shears, silver}
\end{minipage}}\\
\hline

\stepcounter{utterance}\theutterance &
\colorbox[rgb]{1,0.85,0.72}{%
\begin{minipage}{0.3\textwidth}
\texttt{\tiny [GM$\langle$B]}\\
\texttt{GUESS: bluegrass, gospel}
\end{minipage}} 
&
\colorbox[rgb]{1,0.85,0.72}{%
\begin{minipage}{0.3\textwidth}
\texttt{\tiny [GM$\langle$B]}\\
\texttt{GUESS: entrance, china, basement}
\end{minipage}}
&
\colorbox[rgb]{1,0.85,0.72}{%
\begin{minipage}{0.3\textwidth}
\texttt{\tiny [GM$\langle$B]}\\
\texttt{GUESS: attic, entrance, basement}
\end{minipage}} \\
\hline

\stepcounter{utterance}\theutterance &
\colorbox[rgb]{0.95,0.95,0.95}{%
\begin{minipage}{0.3\textwidth}
\texttt{\tiny [A$\langle$GM]}\\
\texttt{The words bluegrass, gospel were guessed correctly. Your teammate's turn ended there.
Now provide another clue relating to some of your remaining team words and a list of the related target words. Remember to start your clue with 'CLUE: ', put a new line, and start your comma-separated list of target words with 'TARGETS: '. Notice: some words have been removed from the lists compared to previous requests.\\
\texttt{}\\
Your remaining team words are: shears, attic, entrance, hose.
Remaining words for your opponent are: head, buttocks, running, back, neck.
Remaining distractor words are: seasoning, brazil, oregano, china, portugal, truck, atv.
Remaining kill words are: silver.}
\end{minipage}}
&
\colorbox[rgb]{0.95,0.95,0.95}{%
\begin{minipage}{0.3\textwidth}
\texttt{\tiny [A$\langle$GM]}\\
\texttt{The word entrance was guessed correctly. The word china was guessed but is an innocent word. Your teammate's turn ended there.
Now provide another clue relating to some of your remaining team words and a list of the related target words. Remember to start your clue with 'CLUE: ', put a new line, and start your comma-separated list of target words with 'TARGETS: '. Notice: some words have been removed from the lists compared to previous requests.\\
\texttt{}\\
Your remaining team words are: gospel, shears, bluegrass, basement, hose.
Remaining words for your opponent are: cotton, buttocks, swimming, neck.
Remaining distractor words are: seasoning, brazil, oregano, portugal, truck, atv.
Remaining kill words are: silver.}
\end{minipage}}
&
\colorbox[rgb]{0.95,0.95,0.95}{%
\begin{minipage}{0.3\textwidth}
\texttt{\tiny [A$\langle$GM]}\\
\texttt{The words attic, entrance, basement were guessed correctly. Your teammate's turn ended there.
Now provide another clue relating to some of your remaining team words and a list of the related target words. Remember to start your clue with 'CLUE: ', put a new line, and start your comma-separated list of target words with 'TARGETS: '. Notice: some words have been removed from the lists compared to previous requests.\\
\texttt{}\\
Your remaining team words are: gospel, shears, hose.
Remaining words for your opponent are: buttocks, running, wool, swimming, neck.
Remaining distractor words are: seasoning, brazil, oregano, china, portugal, truck, atv.
Remaining kill words are: silver.}
\end{minipage}} \\
\hline

\stepcounter{utterance}\theutterance &
\colorbox[rgb]{0.8,1,0.9}{%
\begin{minipage}{0.3\textwidth}
\texttt{\tiny [A$\rangle$GM]}\\
\texttt{CLUE: yard}\\
\texttt{TARGETS: shears, hose}
\end{minipage}} 
&
\colorbox[rgb]{0.8,1,0.9}{%
\begin{minipage}{0.3\textwidth}
\texttt{\tiny [A$\rangle$GM]}\\
\texttt{CLUE: tool}\\
\texttt{TARGETS: shears, hose}
\end{minipage}}
&
\colorbox[rgb]{0.8,1,0.9}{%
\begin{minipage}{0.3\textwidth}
\texttt{\tiny [A$\rangle$GM]}\\
\texttt{CLUE: door}\\
\texttt{TARGETS: entrance, attic}
\end{minipage}}\\
\hline

\stepcounter{utterance}\theutterance &
\colorbox[rgb]{0.95,0.95,0.95}{%
\begin{minipage}{0.3\textwidth}
\texttt{\tiny [GM$\rangle$B]}\\
\texttt{The words bluegrass, gospel were guessed correctly. Your turn ended there.
Now provide another comma-separated list of at least 1 and up to 2 words from the following list of words that best relate or are most closely associated with the word 'door'. Remember to start your answer with 'GUESS: '. Notice: some words have been removed from the list compared to previous requests.\\
\texttt{}\\
attic, running, neck, atv, brazil, portugal, seasoning, entrance, truck, china, buttocks, oregano, hose, shears, back, silver}
\end{minipage}}
&
\colorbox[rgb]{0.95,0.95,0.95}{%
\begin{minipage}{0.3\textwidth}
\texttt{\tiny [GM$\rangle$B]}\\
\texttt{The word entrance was guessed correctly. The word china was guessed but is an innocent word. Your turn ended there.
Now provide another comma-separated list of at least 1 and up to 2 words from the following list of words that best relate or are most closely associated with the word 'tool'. Remember to start your answer with 'GUESS: '. Notice: some words have been removed from the list compared to previous requests.\\
\texttt{}\\
neck, bluegrass, atv, brazil, portugal, seasoning, truck, oregano, swimming, gospel, cotton, basement, hose, shears, silver}
\end{minipage}}
&
\colorbox[rgb]{0.95,0.95,0.95}{%
\begin{minipage}{0.3\textwidth}
\texttt{\tiny [GM$\rangle$B]}\\
\texttt{The words attic, entrance, basement were guessed correctly. Your turn ended there.
Now provide another comma-separated list of at least 1 and up to 2 words from the following list of words that best relate or are most closely associated with the word 'yard'. Remember to start your answer with 'GUESS: '. Notice: some words have been removed from the list compared to previous requests.\\
\texttt{}\\
running, atv, brazil, portugal, seasoning, truck, china, buttocks, oregano, swimming, wool, gospel, hose, shears, silver}
\end{minipage}} \\
\hline

\stepcounter{utterance}\theutterance &
\colorbox[rgb]{1,0.85,0.72}{%
\begin{minipage}{0.3\textwidth}
\texttt{\tiny [GM$\langle$B]}\\
\texttt{GUESS: entrance, door}
\end{minipage}} 
&
\colorbox[rgb]{1,0.85,0.72}{%
\begin{minipage}{0.3\textwidth}
\texttt{\tiny [GM$\langle$B]}\\
\texttt{GUESS: atv, shears}
\end{minipage}} 
&
\colorbox[rgb]{1,0.85,0.72}{%
\begin{minipage}{0.3\textwidth}
\texttt{\tiny [GM$\langle$B]}\\
\texttt{GUESS: atv, hose}
\end{minipage}}  \\
\hline

\stepcounter{utterance}\theutterance &
\colorbox[rgb]{0.95,0.95,0.95}{%
\begin{minipage}{0.3\textwidth}
\texttt{\tiny [GM$|$GM]}
\texttt{Guessed word 'door' is the same word as the provided clue word, you should only select words from the provided list.} \\
\texttt{\tiny [GM$|$GM]}
\texttt{\textbf{Game Aborted}}
\end{minipage}}
&
\colorbox[rgb]{0.95,0.95,0.95}{%
\begin{minipage}{0.3\textwidth}
\texttt{\tiny [A$\langle$GM]}\\
\texttt{The word atv was guessed but is an innocent word. Your teammate's turn ended there.
Now provide another clue relating to some of your remaining team words and a list of the related target words. Remember to start your clue with 'CLUE: ', put a new line, and start your comma-separated list of target words with 'TARGETS: '. Notice: some words have been removed from the lists compared to previous requests.\\
\texttt{}\\
Your remaining team words are: gospel, shears, bluegrass, basement, hose.
Remaining words for your opponent are: cotton, swimming, neck.
Remaining distractor words are: seasoning, brazil, oregano, portugal, truck.
Remaining kill words are: silver.}
\end{minipage}}
&
\colorbox[rgb]{0.95,0.95,0.95}{%
\begin{minipage}{0.3\textwidth}
\texttt{\tiny [A$\langle$GM]}\\
\texttt{The word atv was guessed but is an innocent word. Your teammate's turn ended there.
Now provide another clue relating to some of your remaining team words and a list of the related target words. Remember to start your clue with 'CLUE: ', put a new line, and start your comma-separated list of target words with 'TARGETS: '. Notice: some words have been removed from the lists compared to previous requests.\\
\texttt{}\\
Your remaining team words are: gospel, shears, hose.
Remaining words for your opponent are: buttocks, running, wool, swimming.
Remaining distractor words are: seasoning, brazil, oregano, china, portugal, truck.
Remaining kill words are: silver.}
\end{minipage}} \\
\hline

\stepcounter{utterance}\theutterance &
&
\colorbox[rgb]{0.8,1,0.9}{%
\begin{minipage}{0.3\textwidth}
\texttt{\tiny [A$\rangle$GM]}\\
\texttt{CLUE: music}\\
\texttt{TARGETS: gospel, bluegrass}
\end{minipage}}
&
\colorbox[rgb]{0.8,1,0.9}{%
\begin{minipage}{0.3\textwidth}
\texttt{\tiny [A$\rangle$GM]}\\
\texttt{CLUE: water}\\
\texttt{TARGETS: hose}
\end{minipage}}\\
\hline

\stepcounter{utterance}\theutterance &
&
\colorbox[rgb]{0.95,0.95,0.95}{%
\begin{minipage}{0.3\textwidth}
\texttt{\tiny [GM$\rangle$B]}\\
\texttt{The word atv was guessed but is an innocent word. Your turn ended there.
Now provide another comma-separated list of at least 1 and up to 2 words from the following list of words that best relate or are most closely associated with the word 'music'. Remember to start your answer with 'GUESS: '. Notice: some words have been removed from the list compared to previous requests.\\
\texttt{}\\
neck, bluegrass, brazil, portugal, seasoning, truck, oregano, swimming, gospel, basement, hose, shears, silver}
\end{minipage}}
&
\colorbox[rgb]{0.95,0.95,0.95}{%
\begin{minipage}{0.3\textwidth}
\texttt{\tiny [GM$\rangle$B]}\\
\texttt{The word atv was guessed but is an innocent word. Your turn ended there.
Now provide another comma-separated list of at least 1 and up to 1 words from the following list of words that best relate or are most closely associated with the word 'water'. Remember to start your answer with 'GUESS: '. Notice: some words have been removed from the list compared to previous requests.\\
\texttt{}\\
running, brazil, portugal, seasoning, truck, china, buttocks, oregano, swimming, gospel, hose, shears, silver}
\end{minipage}} \\
\hline

\stepcounter{utterance}\theutterance &
&
\colorbox[rgb]{1,0.85,0.72}{%
\begin{minipage}{0.3\textwidth}
\texttt{\tiny [GM$\langle$B]}\\
\texttt{GUESS: bluegrass, gospel}
\end{minipage}} 
&
\colorbox[rgb]{1,0.85,0.72}{%
\begin{minipage}{0.3\textwidth}
\texttt{\tiny [GM$\langle$B]}\\
\texttt{GUESS: hose}
\end{minipage}}  \\
\hline

\stepcounter{utterance}\theutterance &
&
\colorbox[rgb]{0.95,0.95,0.95}{%
\begin{minipage}{0.3\textwidth}
\texttt{\tiny [A$\langle$GM]}\\
\texttt{The words bluegrass, gospel were guessed correctly. Your teammate's turn ended there.
Now provide another clue relating to some of your remaining team words and a list of the related target words. Remember to start your clue with 'CLUE: ', put a new line, and start your comma-separated list of target words with 'TARGETS: '. Notice: some words have been removed from the lists compared to previous requests.\\
\texttt{}\\
Your remaining team words are: shears, basement, hose.
Remaining words for your opponent are: swimming, neck.
Remaining distractor words are: seasoning, brazil, oregano, portugal, truck.
Remaining kill words are: silver.}
\end{minipage}}
&
\colorbox[rgb]{0.95,0.95,0.95}{%
\begin{minipage}{0.3\textwidth}
\texttt{\tiny [A$\langle$GM]}\\
\texttt{The word hose was guessed correctly. Your teammate's turn ended there.
Now provide another clue relating to some of your remaining team words and a list of the related target words. Remember to start your clue with 'CLUE: ', put a new line, and start your comma-separated list of target words with 'TARGETS: '. Notice: some words have been removed from the lists compared to previous requests.\\
\texttt{}\\
Your remaining team words are: gospel, shears.
Remaining words for your opponent are: buttocks, running, swimming.
Remaining distractor words are: seasoning, brazil, oregano, china, portugal, truck.
Remaining kill words are: silver.}
\end{minipage}}\\
\hline

\stepcounter{utterance}\theutterance &
&
\colorbox[rgb]{0.8,1,0.9}{%
\begin{minipage}{0.3\textwidth}
\texttt{\tiny [A$\rangle$GM]}\\
\texttt{CLUE: storage}\\
\texttt{TARGETS: basement, hose}
\end{minipage}}
&
\colorbox[rgb]{0.8,1,0.9}{%
\begin{minipage}{0.3\textwidth}
\texttt{\tiny [A$\rangle$GM]}\\
\texttt{CLUE: music}\\
\texttt{TARGETS: gospel}
\end{minipage}}\\
\hline

\stepcounter{utterance}\theutterance &
&
\colorbox[rgb]{0.95,0.95,0.95}{%
\begin{minipage}{0.3\textwidth}
\texttt{\tiny [GM$\rangle$B]}\\
\texttt{The words bluegrass, gospel were guessed correctly. Your turn ended there.
Now provide another comma-separated list of at least 1 and up to 2 words from the following list of words that best relate or are most closely associated with the word 'storage'. Remember to start your answer with 'GUESS: '. Notice: some words have been removed from the list compared to previous requests.\\
\texttt{}\\
brazil, portugal, seasoning, truck, oregano, swimming, basement, hose, shears, silver}
\end{minipage}}
&
\colorbox[rgb]{0.95,0.95,0.95}{%
\begin{minipage}{0.3\textwidth}
\texttt{\tiny [GM$\rangle$B]}\\
\texttt{The word hose was guessed correctly. Your turn ended there.
Now provide another comma-separated list of at least 1 and up to 1 words from the following list of words that best relate or are most closely associated with the word 'music'. Remember to start your answer with 'GUESS: '. Notice: some words have been removed from the list compared to previous requests.\\
\texttt{}\\
running, brazil, portugal, seasoning, truck, china, buttocks, oregano, gospel, shears, silver}
\end{minipage}} \\
\hline

\stepcounter{utterance}\theutterance &
&
\colorbox[rgb]{1,0.85,0.72}{%
\begin{minipage}{0.3\textwidth}
\texttt{\tiny [GM$\langle$B]}\\
\texttt{GUESS: basement, silver}
\end{minipage}} 
&
\colorbox[rgb]{1,0.85,0.72}{%
\begin{minipage}{0.3\textwidth}
\texttt{\tiny [GM$\langle$B]}\\
\texttt{GUESS: gospel}
\end{minipage}}  \\
\hline

\stepcounter{utterance}\theutterance &
&
\colorbox[rgb]{0.95,0.95,0.95}{%
\begin{minipage}{0.3\textwidth}
\texttt{\tiny [GM$|$GM]}
\texttt{Team revealed.} \\
\texttt{\tiny [GM$|$GM]}
\texttt{\textbf{Lose}}
\end{minipage}}
&
\colorbox[rgb]{0.95,0.95,0.95}{%
\begin{minipage}{0.3\textwidth}
\texttt{\tiny [A$\langle$GM]}\\
\texttt{The word gospel was guessed correctly. Your teammate's turn ended there.
Now provide another clue relating to some of your remaining team words and a list of the related target words. Remember to start your clue with 'CLUE: ', put a new line, and start your comma-separated list of target words with 'TARGETS: '. Notice: some words have been removed from the lists compared to previous requests.\\
\texttt{}\\
Your remaining team words are: shears.
Remaining words for your opponent are: buttocks, running.
Remaining distractor words are: seasoning, brazil, oregano, china, portugal, truck.
Remaining kill words are: silver.}
\end{minipage}}\\
\hline

\stepcounter{utterance}\theutterance &
&
&
\colorbox[rgb]{0.8,1,0.9}{%
\begin{minipage}{0.3\textwidth}
\texttt{\tiny [A$\rangle$GM]}\\
\texttt{CLUE: garden}\\
\texttt{TARGETS: shears}
\end{minipage}}\\
\hline

\stepcounter{utterance}\theutterance &
&
&
\colorbox[rgb]{0.95,0.95,0.95}{%
\begin{minipage}{0.3\textwidth}
\texttt{\tiny [GM$\rangle$B]}\\
\texttt{The word gospel was guessed correctly. Your turn ended there.
Now provide another comma-separated list of at least 1 and up to 1 words from the following list of words that best relate or are most closely associated with the word 'garden'. Remember to start your answer with 'GUESS: '. Notice: some words have been removed from the list compared to previous requests.\\
\texttt{}\\
brazil, portugal, seasoning, truck, china, buttocks, oregano, shears, silver}
\end{minipage}} \\
\hline

\stepcounter{utterance}\theutterance &
&
&
\colorbox[rgb]{1,0.85,0.72}{%
\begin{minipage}{0.3\textwidth}
\texttt{\tiny [GM$\langle$B]}\\
\texttt{GUESS: shears}
\end{minipage}}  \\
\hline

\stepcounter{utterance}\theutterance &
&
&
\colorbox[rgb]{0.95,0.95,0.95}{%
\begin{minipage}{0.3\textwidth}
\texttt{\tiny [GM$|$GM]}
\texttt{\textbf{Game Success}}
\end{minipage}} \\
\hline

\end{longtable}

\twocolumn





\end{document}